\documentclass[10pt,twocolumn,letterpaper]{article}

\usepackage{standalone}
\usepackage{iccv}
\usepackage{times}
\usepackage{epsfig}
\usepackage{graphicx}
\usepackage{amsmath}
\usepackage{amssymb}

\usepackage{color}
\usepackage{colortbl}
\usepackage[table]{xcolor}
\usepackage{import}
\usepackage[accsupp]{axessibility}
\usepackage{booktabs}       % professional-quality tables
\usepackage{amsfonts}       % blackboard math symbols
\usepackage{nicefrac}       % compact symbols for 1/2, etc.
\usepackage{microtype}      % microtypography
\usepackage{pifont}% http://ctan.org/pkg/pifont
\usepackage[ruled,vlined]{algorithm2e} %algorithms
\usepackage[font=small,skip=2pt]{caption}
\usepackage[inline]{enumitem}

\usepackage{cuted}
\usepackage{capt-of}

\newcommand{\rcomment}[1]{}{}%{#1}
\newcommand{\amirg}[1]{\textcolor{red}{\rcomment{Amir: #1}}}

\newcommand{\phillip}[1]{\textcolor{blue}{\rcomment{Phillip: #1}}}

\newcommand{\gal}[1]{\textcolor{magenta}{\rcomment{Gal: #1}}}
\newcommand{\yossig}[1]{\textcolor{orange}{\rcomment{Yossi: #1}}}
\newcommand{\oran}[1]{\textcolor{brown}{\rcomment{Oran: #1}}}

\newcommand{\inbarm}[1]{\textcolor{purple}{\rcomment{Inbar: #1}}}

\newcommand{\sspacename}{Wu {\em et al.}}

\newcommand{\ours}{StylEx}

\newcommand\blfootnote[1]{%
  \begingroup
  \renewcommand\thefootnote{}\footnote{#1}%
  \addtocounter{footnote}{-1}%
  \endgroup
}
\pdfoutput=1

\newcommand{\x}{\mathbf{x}}
\newcommand{\comment}[1]{}
\newcommand{\secref}[1]{Sec.~\ref{#1}}
\newcommand{\figref}[1]{Fig.~\ref{#1}}
\newcommand{\algref}[1]{Alg.~\ref{#1}}
\newcommand{\tabref}[1]{Table~\ref{#1}}

\usepackage{hyperref}
\hypersetup{pagebackref=true,breaklinks=true,letterpaper=true,colorlinks,bookmarks=false}

\iccvfinalcopy % *** Uncomment this line for the final submission

\def\iccvPaperID{7751} % *** Enter the ICCV Paper ID here

\ificcvfinal\fi

\begin{document}

%%%%%%%%% TITLE
%\title{Generative Explanations: Training a Generator to Visually Explain a Classifier}
% \title{GenX: Attribute-Based Generative Explanations of Classifiers}
%\title{Explaining in Style: A StyleGAN Trained to Explain a Classifier }
\title{\vspace*{-0.5cm}Explaining in Style: Training a GAN to explain a classifier in StyleSpace}
% \and

\author{Oran Lang$^{*1}$
\qquad
Yossi Gandelsman$^{*1}$
\qquad
Michal Yarom$^{*1}$
\qquad
Yoav Wald$^{*1,2}$
\\
Gal Elidan$^{1}$
\qquad
Avinatan Hassidim$^{1}$
\qquad
William T. Freeman$^{1,3}$
\qquad
Phillip Isola$^{1,3}$
\\
Amir Globerson$^{1,4}$
\qquad
Michal Irani$^{1,5}$
\qquad
Inbar Mosseri$^1$ \vspace{0.2cm}
\\
\small $^1$ Google Research \qquad $^2$ Hebrew University\qquad $^3$ MIT \qquad $^4$ Tel Aviv University\qquad $^5$Weizmann Institute of Science
}

\maketitle

\ificcvfinal\thispagestyle{empty}\fi

\begin{strip}
\vspace{-0.6in}
\centering
% \vspace{-10cm}
\begin{minipage}{\textwidth}
% \vspace{-3cm}
	\centering
\includegraphics[width=0.95\textwidth]{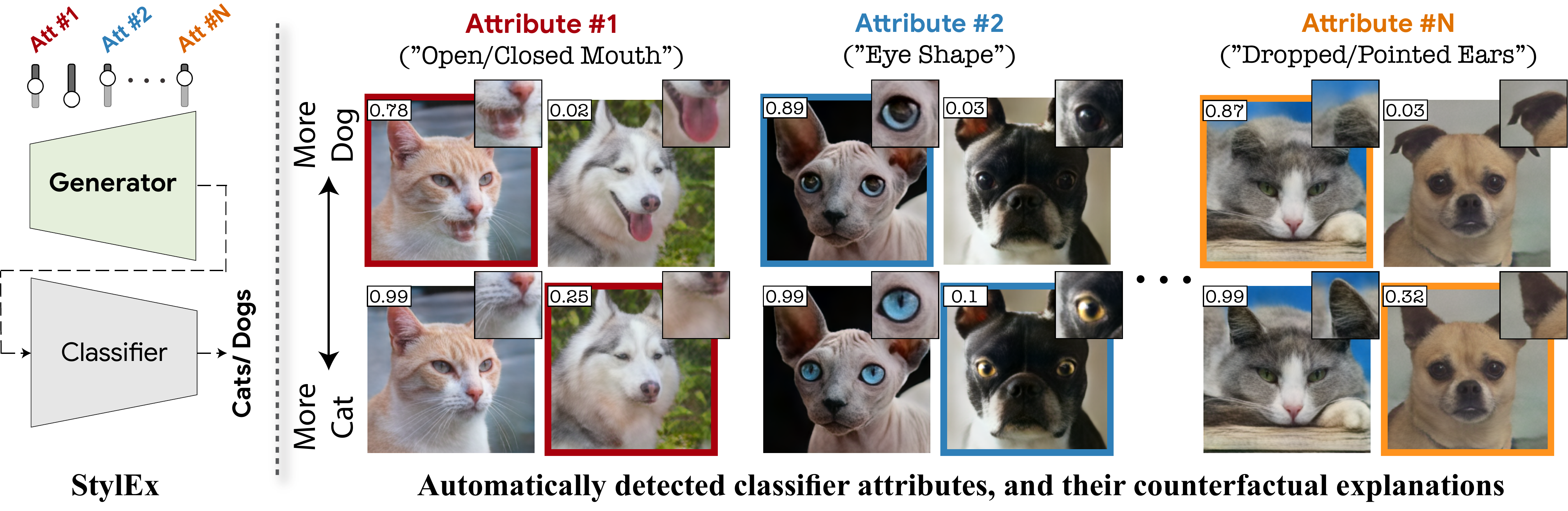}
%\vspace{-0.1cm}
\captionof{figure}
% {\textbf{Classifier-specific interpretable attributes emerge in the \ours{} StyleSpace.} Our system, \ours, explains the decisions of a classifier by discovering and visualizing multiple attributes that affect its output. (Left) StylEx achieves this by training a StyleGAN specifically to explain the classifier, discovering latent attributes in the GAN's StyleSpace that affect the classifer's decisions. (Right)~Applied to a ``cat vs dog" classifier, the system automatically discovers key visual attributes corresponding to StyleSpace coordinates, and
% visualizes how manipulating these coordinates affects the classifier output probability (shown in the top-left of each image). The top attributes found by our method indeed correspond to coherent semantic properties that affect perception of cats vs dogs  (e.g. open or closed mouth, eye shape, and pointed or dropped ears).
% }
{\textbf{Classifier-specific interpretable attributes emerge in the \ours{} StyleSpace.}
Our system, \ours, explains the decisions of a classifier by discovering and visualizing multiple attributes that affect its prediction. (Left) StylEx achieves this by training a StyleGAN specifically to explain the classifier (e.g., a ``cat vs. dog" classifier), thus driving latent attributes in the GAN's StyleSpace to capture \emph{classifier-specific} attributes. (Right)~We automatically discover top visual attributes in the StyleSpace coordinates, which best explain the classifier's decision. For each discovered attribute, \ours{} can then provide an explanation by generating a \emph{counterfactual example}, i.e., visualizing how manipulation of this attribute (style coordinate) affects the classifier output probability. The generated counterfactual examples are marked in the figure as the images with colored borders.\amirg{add: The other images are real. Alternatively write: Each column shows an original image and a ``counter-factual'' image generated by StyleEx where a single attribute is changed.} The degree to which manipulating each attribute affects the classifier probability is shown in the top-left of each image. The top attributes found by our method indeed correspond to coherent semantic properties that affect perception of cats vs. dogs  (e.g. open or closed mouth, eye shape, and pointed or dropped ears).
}
%(names were provided manually after inspecting multiple modifications). 
%\yossig{I think we should add the fact that the discovered attributes are unsupervised, and the naming is something we create.}}
%\ours  achieves this by training a StyleGAN guided by the classifier to discover latent attributes, in the GAN's StyleSpace, that most affect the classifier's decisions.}
%Classifier specific disentangled attributes emerge in the StylEX style-space when trained with the classifier}
% \afterfig
\label{fig:teaser}
\end{minipage}
\end{strip}

%%%%%%%%% ABSTRACT
\blfootnote{$^*$ Equal contributors; Work performed by authors at Google.}

\begin{abstract}
%Explaining the decision of visual classifiers is an important 
Image classification models can depend on multiple different semantic attributes of the image. An explanation of the decision of the classifier needs to both discover and visualize these properties. Here we present {\ours}, a method for doing this, by training a generative model to specifically explain multiple attributes that underlie classifier decisions. A natural source for such attributes is the StyleSpace of StyleGAN, which is known to generate semantically meaningful dimensions in the image. However, because standard GAN training is not dependent on the classifier, it may not represent those attributes which are important for the classifier decision, and the dimensions of StyleSpace may represent irrelevant attributes. 
%However, these will typically not correspond to classifier-specific attributes since standard GAN training is not dependent on the classifier. 
% To overcome this, we propose a training procedure for
% a StyleGAN, which incorporates the classifier model, in order to learn a classifier-specific StyleSpace. Explanatory attributes are then selected from this space.
To overcome this, we propose a training procedure for a StyleGAN, which incorporates the classifier model, in order to learn a classifier-specific StyleSpace. Explanatory attributes are then selected from this space.
%\oran{This is not accurate.  The training procedure is a solution to the GAN not learning to represent the semantic attributes of the classifier well. I suggest:
%However, because standard GAN training is not dependent on the classifier, it may not represent these attributes which are important for the classifier decision, and many dimensions of StyleSpace... }
%This results in an S-space that captures distinct attributes underlying classifier outputs. 
%We introduce a procedure for discovering coordinates of the trained StyleSpace whose modification affects classifier output.

%Here we present a method for discovering and visualizing these properties.
These can be used to visualize the effect of changing multiple attributes per image, thus providing image-specific explanations. We apply StylEx to multiple domains, including animals, leaves, faces and retinal images. For these, we show how an image can be modified in different ways to change its classifier output.
Our results show that the method finds attributes that align well with semantic ones, generate meaningful image-specific explanations, and are human-interpretable as measured in user-studies.\footnote{\bf Project website: \ \scalebox{0.91}[1.0]{\bf \url{https://explaining-in-style.github.io/}}}
\end{abstract}

\begin{figure*}[t]
   \centering
   \includegraphics[width=\linewidth]{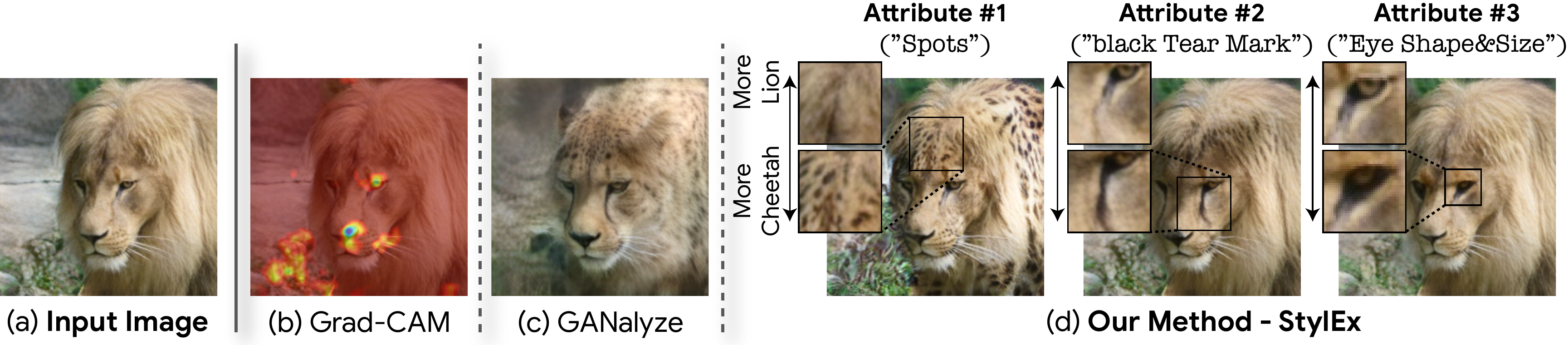}
   \vspace{-.2in}
   \caption{{\bf Comparison to other visual explanation methods for a Lion vs. Cheetah classifier}. (b) Grad-CAM \cite{selvaraju2017grad} and other heat-map based methods are limited in their ability to visualize attributes that are not spatially localized (e.g., eye size). %do not provide sufficient semantic information about the classification decision%
   (c) GANalyze \cite{goetschalckx2019ganalyze} produces a possible counterfactual explanation, but its visualization changes all relevant attributes at once. (d) Our StylEx method provides meaningful interpertable multi-attribute explanation, by generating counterfactuals that change one attribute at a time.}
   \label{fig:motivation}
   \vspace*{-0.5cm}
\end{figure*}
\vspace{-0.2in}
\section{Introduction}
%\amirg{There is probably other work than Lichinski that shows use of S space. If we find these, let's mention throughout}
%Explanations of image classification models are beneficial for a range of purposes from detecting model biases to providing support to human decision makers and aiding scientific discovery \cite{narayanaswamy2020scientific}. 
%higher-level semantic attributes (e.g., facial-hair). However, since these attributes are not known a-priori, the key challenge is to discover them and build mechanisms for changing them in a given image. This is the problem we address. % in the current work.
Deep net classifiers are often described as ``black boxes" whose decisions are opaque and hard for humans to understand. This significantly holds back their usage, especially in areas of high societal impact, such as medical imaging and autonomous driving, where human oversight is critical. Explaining the decision of classifiers can reveal model biases \cite{kim2018interpretability}, provide support to downstream human decision makers, and even aid scientific discovery~\cite{narayanaswamy2020scientific}. 

Among the different forms of explanations, {counterfactual explanations} are gaining
% \amirg{I don't think we should say based on generative mmodes. That's not common and is part of our contribution.}\yoavw{I agree and also the references are not for generative methods. Someone must have added ``generative" recently, removing it now but if someone prefers it this way then please tell me.}
increasing attention \cite{mothilal2020explaining, goyal2019counterfactual, wachter2017counterfactual}. A \emph{counterfactual explanation} is a statement of the form ``Had the input $\mathbf{x}$ been $\tilde{\mathbf{x}}$ then the classifier output would have been $\tilde{y}$ instead of $y$", where the difference between $\mathbf{x}$ and $\tilde{\mathbf{x}}$ is easy to explain. For instance, consider a classifier trained to distinguish between cat and dog images. A counterfactual explanation for an image classified as a cat could be ``If the pupils were made larger, then the output of the classifier for the probability of cat would decrease by 10\%."
A key advantage of this approach is that it provides per-example explanations, pinpointing which parts of the input are salient towards the classification and also how they can be changed in order to obtain an alternative outcome. 
% \gal{At some point, not sure if here, above in the references or in related works, we want to make the distinction between explaining the model in general and per-example explanations more concrete by also including references of the other types}

The effectiveness of counterfactual explanations strongly depends on how intuitive  the difference between $\mathbf{x}$ and $\tilde{\mathbf{x}}$ is to human observers. For instance, if $\tilde{\mathbf{x}}$ is an arbitrary dog image,  it is not useful as a counterfactual explanation since it usually changes all features of $\x$, hence does not isolate the critical features the classifier depends on. Adversarial examples \cite{goodfellow2014explaining,guo2019simple} , which are slight modifications to the input $\mathbf{x}$ that change the classification to the wrong class, are also not effective counterfactual explanations, as the changes are usually not interpretable by humans.

Therefore to form a useful counterfactual explanation we must discover interpretable features, or \emph{attributes}; in the case of the cats vs. dogs classifier, these might be ``pupil size" or ``open mouth''.
%, such as pupil size and open mouth. 
To visualize them we further need to enable control of these attributes in the image, a task most suited to generative models. This is an inherently different task from visualizing a smooth transformation between one class and the other, as done for instance in \cite{goetschalckx2019ganalyze, singla2019explanation}. Such transformations change all attributes at once, making it difficult to isolate fine grained attributes. 
% Figure \ref{fig:teaser} shows that our approach \ours{}, outlined below, indeed succeeds in capturing and visualizing these attributes.% in this case. 
% \gal{Despite adding 'outlined below', this sentence still feel a bit out of place. Consider leading up to it and moving toward the end of the intro. If we want to refer to the teaser also, we can do that by referring to the figure following one of the counterfactual examples}
Both defining and visualizing interpretable attributes are challenging tasks since in many domains 
(e.g. medical imaging) we may not know the salient visual attributes, or do not have labeled examples of them. 
%Indeed, training generative models that learn a set of disentangled attributes must be guided by some type of supervision or inductive bias \cite{locatello2019challenging}. \gal{Wording here creates a feeling that problem was solved by \cite{locatello2019challenging}}

%\amirg{should be careful with this statement since it's probably impossible also for classification} \phillip{I agree we should be careful: StyleGAN works, yet it is unsupervised and I wouldn't really say it uses ``prior knowledge" (although it does have inductive biases)} \yoavw{Right, inductive biases looks like the right term here. Is the current form of the sentence ``soft" enough, or should we just remove it?}\phillip{I think it's good now and we can keep it.}

%Semantic manipulation of images has been a focus of much recent study, in particular of StyleGAN and its variants. However, these methods seek disentangled directions in the latent space of the GAN, but not attributes that underlie classification decisions. In order to find these attributes, we propose the \ours method, which trains the GAN while taking the classifier into account. Or in other words, training the GAN to explain the classifier. 
A natural candidate for finding visual attributes and visualizing them is generative models, such as StyleGAN2~\cite{karras2020analyzing}, where it was shown that it is possible to find disentangled latent variables that control semantic attributes of the images they generate. This approach has been used to create powerful interfaces for image editing and data visualization 
%\phillip{[citations]}\yoavw{added citations, not sure these are the best ones since I'm not that familiar with the literature}\phillip{these look good, adding ganalyze here too}
\cite{collins2020editing, shen2020interpreting, goetschalckx2019ganalyze, shen2020closed, shen2020interfacegan, Yinghao2020}. Our method builds upon a recent observation by \cite{wu2020stylespace} that StyleGAN2 tends to contain a \emph{disentangled} latent space (i.e., the ``StyleSpace") which can be used to extract individual attributes.
 However, as we show here, this approach will not necessarily discover classifier-related attributes since standard StyleGAN2 training does not involve the classifier. Instead, we propose our \ours{} framework to overcome this difficulty and promote classifier explainability by: (i)~incorporating the classifier into the StyleGAN training procedure to obtain a \mbox{classifier-specific} StyleSpace, and then (ii)~mining this StyleSpace for a concise set of attributes that affect the classifier prediction.

Adding the classifier into the training process of the GAN turns out to be crucial in domains where the classification depends on fine details (e.g., in retinal fundus images). A generator unaware of the importance of these subtle details, may fail to generate them.
% and images of diseased leaves.
%Adding the classifier into the training process turns out to be crucial in order to learn attributes that affect the classifier, especially in domains where the classification depends on fine details such as retinal fundus or leaves images. 
%

% \amircr{
% Our \ours{} method is designed to discover visual attributes that affect the classifier decision. It should be emphasized that we do not argue that these attributes are correlated with the true label. For example, they could correspond to biases of the classifier that result from biases in the training set. Indeed, one potential use of \ours{} is to discover and mitigate classifier bias. 
% }

It should be emphasized that our goal is not to explain the  true label, but rather what classifiers are learning. For example, they could correspond to biases of the classifier that result from biases in the training set. Indeed, one potential use of \ours{} is to discover and mitigate classifier biases.

We demonstrate \ours{} on a variety of domains, 
%where StyleGANs are successful, 
and show it extracts semantic attributes that are salient for classification in each domain.

%\phillip{contribution list needs work, or could remove}\yoavw{I think Amir added this and I also think it's good to have a paragraph that summarizes contributions, since otherwise readers may get confused about what is the actual novelty here. Maybe it'll be best to revise these when most results are in place?}\phillip{I think that's a good plan. Let's come back to it.}

%\vspace*{0.2cm}
\smallskip\noindent
Our contributions are as follows: 
\begin{itemize}[topsep=0pt,itemsep=-1ex,partopsep=1ex,parsep=1ex]
%     \item We propose the setup of \textit{multi-attribute} counterfactual explanations of vision models.
%     %\item We show how such models can be trained by training a GAN using a pre-trained classifier we want to explain.
%     \item We propose the \ours{} model for classifier-based training of a StyleGAN2 towards finding and visualizing attributes.
%     \item We evaluate the method on multiple domains, showing it provides explanations that are understood by human users.
% \end{itemize}
%
    %\item {$\bullet$} A general setup of %\textit{multi-attribute} counterfactual explanations of image-based classifiers.
    %\item We show how such models can be trained by training a GAN using a pre-trained classifier we want to explain.
\item We propose the \ours{} model for classifier-based training of a StyleGAN2, thus driving its StyleSpace to capture \emph{classifier-specific} attributes.
    %towards finding and visualizing classifier-specific attributes.
\item A method to discover classifier-related attributes in StyleSpace coordinates, and use these for counterfactual explanations.
    %which best explain a classifier's decision. 
\item \ours{} is applicable for explaining a large variety of classifiers and real-world complex domains. We show it provides explanations understood by human users.
\end{itemize}

%\gal{The difference between point \#1 and point \#2 is not crisp. And, we are missing the point of training the explanation space \emph{with} the classifier. I suggest making \#1: learning a counterfactual space, \#2,\#3: as they are now}
\comment{
However, recent advances in generative models show that it is often possible to extract and manipulate semantic attributes from latent representations by using carefully designed architectures and training methods. Architectures like StyleGAN \cite{karras2019style} and methods such as those proposed in \cite{peebles2020hessian,harkonen2020ganspace, plumerault2020controlling, voynov2020unsupervised} demonstrate increasingly improved abilities in extraction of semantic attributes from latent representations of generative models. Recently, \cite{wu2020stylespace} observed surprising disentanglement properties of the so called StyleSpace. Following these works, it is tempting to adapt such methods for the purpose of explaining classifiers.  

In this work we propose \ours{}, a framework for training generative models that are geared towards counterfactual explanations of a given classifier, using multiple attributes learned in an unsupervised manner. \ours{} builds upon the disentanglement properties of Style-GAN generators and uses StyleSpace as a library of semantic attributes. These attributes are then edited in order to create a diverse set of counterfactual explanations. To learn a generator that is aimed at explaining a given classifier, \ours{} includes an encoder which enables the explanation of any image and uses a loss that ensures decoded images retain the classifications of their source images. This proves crucial in order to learn attributes that affect the classifier, especially in domains where the classification depends on fine details such as retinal fundus images (TODO: remove in case we do not include retina data). We demonstrate \ours on a variety of domains where Style-GANs are successful, and show it extracts semantic attributes that are salient for classification in each of the domains.

We next describe in more detail how \ours explains classifications and how it relates to other proposed forms of explanations.
}
% Figures: Architecture + visual capabilities
% Domains + top 5 attributes for the classifier
% Top 5 attributes for a certain image.
% Paths to change the classification of an image.

% Quantitative comparison to other methods.
% Accuracy, how many features, diversity, triplet study/,
% Presentation of prior work - slide for each work + results/figure 1s.
\section{Related Work}
\label{sec:related}
% \begin{table*}[htp!]
%     \centering
%     \scalebox{0.8}{
%     \begin{tabular}{|l|ccc|ccc|} \toprule
%         & \multicolumn{3}{c|}{Visualization} & \multicolumn{3}{c}{Attributes/Concept} \\ 
%         & Generative Model & Heatmaps & Examplars & Raw Features & Predefined & Extracted \\ \midrule \midrule
%         \cite{singla2019explanation} & \checkmark & & & & & \\
%         \cite{yeh2020completeness, ghorbani2019towards} & & &\checkmark & & & \checkmark \\
%          \cite{selvaraju2017grad} & & \checkmark & & \checkmark & & \\
%          \cite{goyal2019counterfactual} & & \checkmark & & \checkmark & &\\
%          \cite{goyal2019explaining} & \checkmark & & & & \checkmark & \\
%          \bottomrule
%     \end{tabular}}
%     \caption{Visual Explanation Techniques Categorized by Characteristics.}
%     \label{tab:wilds}
%     \vspace{-0.3cm}
% \end{table*}
% \amirg{add this https://genforce.github.io/sefa/}
% \yoavw{added in intro}
% {\bf Add paper on counterfactual visual explanations}
The most widespread visual explanation methods are based on heatmaps. Such maps highlight regions of the image that are salient towards the decision, or towards the activation of a hidden unit of the classifier (e.g., \cite{selvaraju2017grad, rebuffi2020there, xu2020attribution}). Heatmaps are useful to understand things like which objects in an image contributed to a classification. However, they cannot visualize/explain well attributes that are not spatially localized, like size, color, etc. In addition, they can show which areas of the image may be changed in order to affect the classification, but not \emph{how} they should be changed.
% \gal{This last sentence shifts the focus back to how to change and not to the problem if being limited to a spatial. Change 'In general' to 'In addition'.}

Counterfactual explanations address these limitations by providing alternative inputs, where a small set of attributes is changed and the different classification outcomes are observed.
% Such explanations are only useful if the attributes are interpretable to humans, unlike adversarial attacks such as \cite{guo2019simple} where a small set of pixels is changed to alter the outcomes. Therefore, counterfactual explanations in computer vision should rely on something other than raw pixels to furnish a set of semantic attributes.
Generative models are natural candidates to produce visual counterfactual explanations, and indeed recent works have shown progress towards this goal. In \cite{singla2019explanation, goetschalckx2019ganalyze, singla2021explaining, antoran2020getting} generative counterfactual explanations are produced, yet their visualization changes all relevant attributes at once, as shown in \figref{fig:motivation}. Another related approach offered in \cite{shocher2020semantic} is to use deep representations from a classifier to manipulate generated images at different granularities. Yet these may involve properties that do not affect the classification outcome and also combine several attributes. Hence these methods do not allow the analysis of a classifier in terms of atomic attributes and their effect on classifications. Other explanation methods generate counterfactuals using attributes, where full or partial supervision for the desired attributes is available \cite{goyal2019explaining, esser2020disentangling}. \cite{goyal2019counterfactual} propose a counterfactual explanation method that is not based on a generative model, and instead replaces a small number of patches from one image into another. Their method does decompose the counterfactual generation into several patch replacements, though the counterfactuals are often not realistic images and the method does not explicitly define a set of controllable attributes. The closest method in spirit to the explanations provided by our \ours{} approach is \cite{o2020generative}, though their method only works on small images, and they do not demonstrate explanations that consist of more than a single change of attributes (nor do they claim to find multiple semantic attributes that affect the classification).

Explanations based on multiple attributes that are extracted in an unsupervised manner are given in \cite{yeh2020completeness, ghorbani2019towards}. They extract attributes based on superpixels, or activations in the mid-layers of the classifier, and do not use a generative model. Hence they do not create images that serve as counterfactuals, and their attributes are demonstrated by showing relevant image patches or superpixels.
% Generating counterfactual images as done in \ours{} is a desirable property when giving explanations based on multiple attributes, especially when the explanation consists of changes in several attributes.
\phillip{the previous sentence does not flow smoothly to the subsequent explanation.}
% \phillip{why is this desirable, can we be more concrete}. \yoavw{The following is an attempt to address Phillip's comment, would be happy to hear your opinion:}
In terms of visualization, representative patches are limited in their ability to visualize attributes that are not spatially localized. Furthermore, counterfactual images let us observe how the classification changes under interventions on a combination of attributes.
% , showing which ones are salient towards individual predictions, and enabling the quantification of salience using their causal effect as advocated by \cite{goyal2019explaining}.
% The discovery and manipulation of attributes provided by \ours{} is made possible by training our GAN to explicitly explain the classifier, as now turn to describe.
\gal{Make this a bit crisper and more explicit - the crux of our work is the \emph{joint training} - we need to hammer that message in}
% \yoavw{consider adding in case we succeed in showing diverse explanations: Another attractive property of \ours{} is the ability to provide multiple diverse counterfactual explanations. This is common in papers on counterfactual explanations, but not on image data nor with attributes discovered in an unsupervised manner (e.g. \cite{dandl2020multi}). Several generative models offer diverse transitions between domains, but not based on attributes nor with the goal of explaining a classifier in mind \cite{zhu2017toward, huang2018multimodal, choi2020stargan}.}
\begin{figure}[t]
   \centering
   \includegraphics[width=\linewidth]{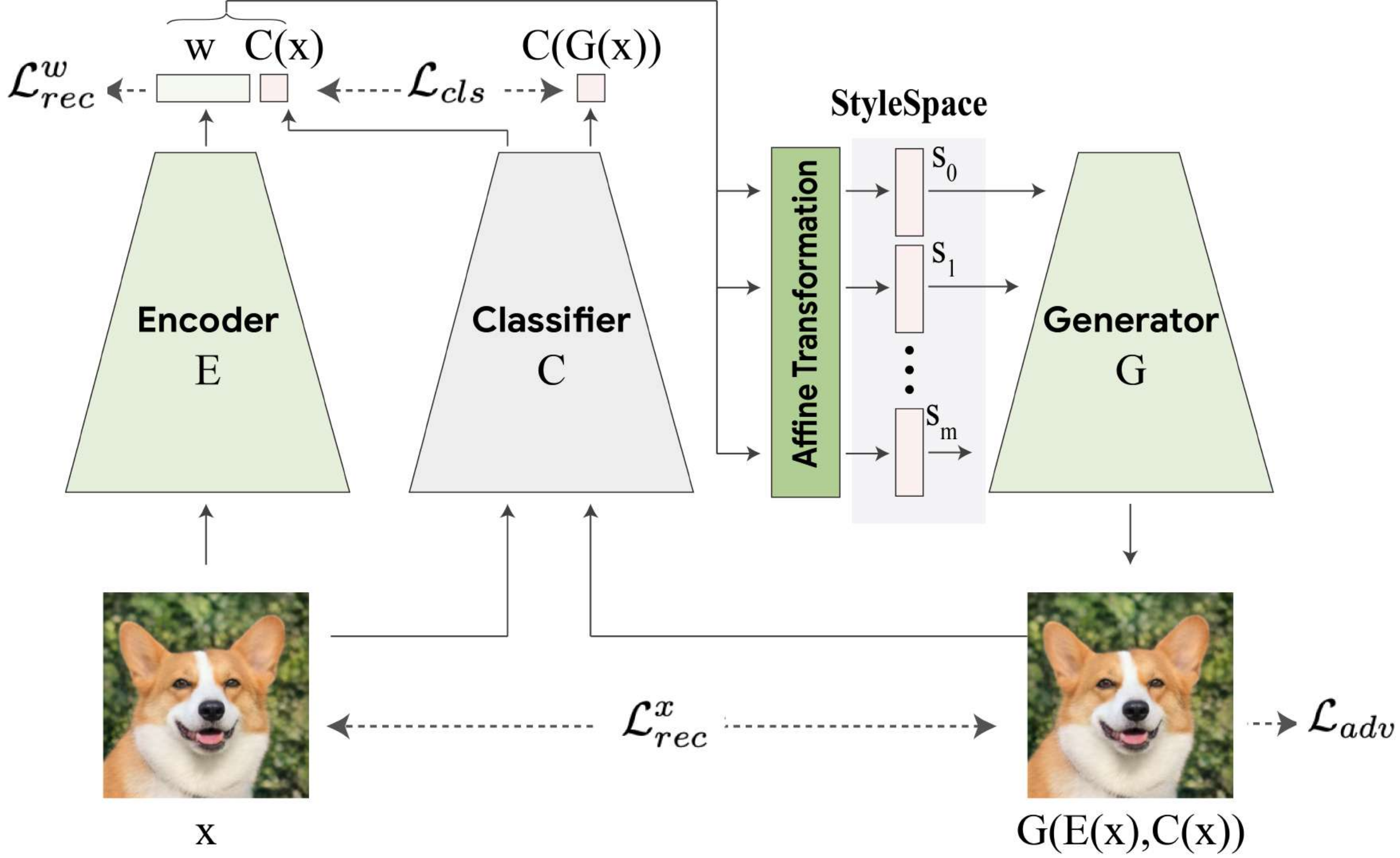}
   \vspace{-.1in}
   \caption{{\bf {\ours} architecture}. We jointly train the generator $G$, discriminator $D$, and encoder $E$. During the training phase, an input image is transformed via the encoder into a latent vector $w$. $w$ is then concatenated to the output $C(x)$ of the classifier $C$ on the image $x$. The result is transformed via affine transformations to the style vectors $s_0, ..., s_n$, which are then used to generate an image close to the original image. A reconstruction loss is applied between the generated image and the original image, as well as between the corresponding encoder outputs. A GAN loss is applied on the generated image, and a KL loss is applied between the output of the classifier $C$ on the generated image and the input condition. 
   %During the inference phase (b), to manipulate a specific attribute in a given image, we generate its style vectors $s_0, ... s_n$, and then change the value of a specific StyleSpace coordinate before passing them to the generator.}
   }
   \label{fig:architecture}
\end{figure}
\begin{figure*}[t]
\vspace*{-0.5cm}
   \centering
   \includegraphics[width=\linewidth]{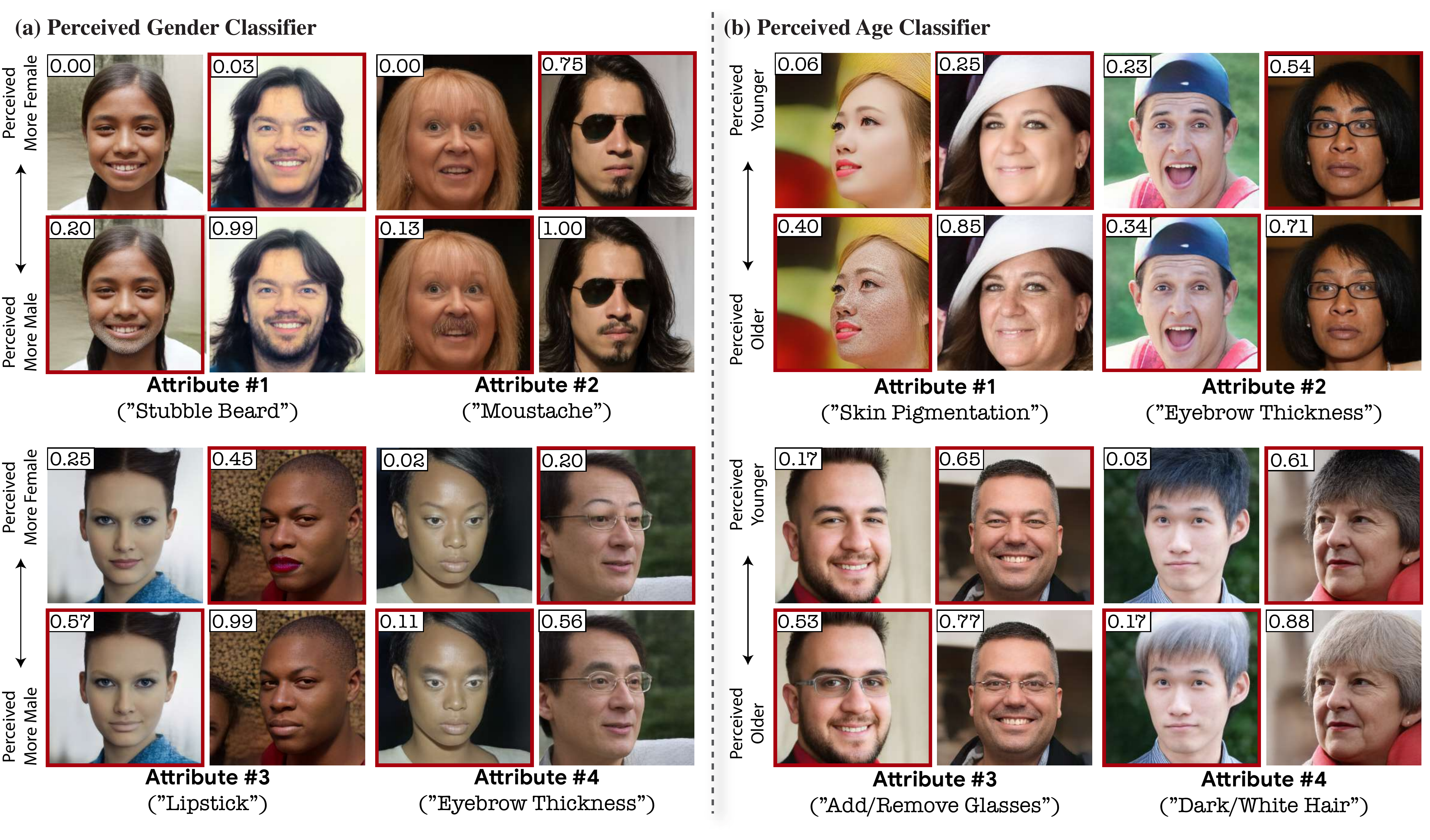}
%   \vspace{-.2in}
   \caption{{\bf Top-4 automatically detected attributes for perceived-gender and perceived-age classifiers}. The corresponding modifications are visually coherent between different images, represent diverse semantic attributes, and affect the classifiers' predictions (presented in the top-left corner of each image) towards the wanted directions. The generated counterfactual examples are marked by a frame. Please refer to the \href{https://explaining-in-style.github.io/supmat.html}{project website results page} for more attributes and animated-GIFs to view these counterfactual changes (explanations) dynamically.}
   \label{fig:gender_top5_attriubtes}
   \vspace*{-0.3cm}
\end{figure*}

\section{Method}
We next describe our approach for discovering classifier-related attributes and modifying these attributes in real images. Our approach consists of two key steps. First, we train a StyleGAN model in a way that incorporates the classifier, thus encouraging the StyleSpace of the StyleGAN to accommodate classifier-specific attributes (\secref{subsection:training_expgan}). Then, we  search the StyleSpace of the trained GAN to automatically discover coordinates that correspond to classifier-specific attributes (\secref{sub:discovering_attributes}). Finally, we show how to use these attributes in order to visually explain a classifier's decision for a given input image (\secref{sub:visualizing_image}).

\subsection{\ours{} Architecture}
%\gal{I don't like this 'based on ' opening and generally, as well as in the title, we appear as some add on to styleGAN. Instead, we should talk in terms of StyleGAN being a component of our approach. I suggest starting with the high level abstract components that we need. Then say something like 'For the generative component, we rely on StyleGAN since ...'}
Recall that our goal is to explain the classification of a given image by changing certain attributes in the image, and to show they affect the classifier output. We achieve this by combining the following components: 
a)~A conditional generative model that maps an embedding $w$ into an output image. b)~An encoder that maps any input image into an embedding $w$, so that the generator can modify attributes in real images. c)~A mechanism for ``intervening'' with the generation process to change visual attributes in the image. 

For the generative model we use StyleGAN2~\cite{karras2020analyzing}. This architecture was shown to generate realistic images in multiple domains. But more important to our goal is the observation recently made by \cite{wu2020stylespace} that  StyleGAN2 tends to inherently contain a disentangled StyleSpace space, which can be used to extract individual attributes. Thus, we argue that modifying coordinates of StyleSpace is a natural approach to our problem of modifying classifier-related attributes.
In~\cite{wu2020stylespace} the authors extracted coordinates of StyleSpace that corresponded to \emph{known attributes} in a pre-trained StyleGAN2. In general, however, StyleGAN2 is not trained to discover classifier-related attributes of an arbitrary classifier, since standard StyleGAN training does not involve that classifier in any way (as shown in \secref{sub:effect_on_classifier}).
%\oran{Suggested edit: However, as we demonstrate in Sec.\ref{sub:effect_on_classifier}, a GAN trained using the standard approach may omit important attributes which the classifier observe, specifically in domains where these attributes include fine details in the image}

%In this paper we show how the inherent property of a StyleGAN2-like architecture, which was shown to give rise to disentangled-attributes, can be harnessed to automatically \emph{discover unknown attributes} of \emph{any classifier of interest} and modify these. 

To overcome the above problem and allow for the StyleSpace to contain classifier-related attributes, we train our GAN to explicitly explain the classifier, by conditioning the model on classifier output and using a classifier loss (see below). As we shall see this will result in \underline{Classifier-specific} disentangled attributes that emerge in our StyleSpace. 

%In StyleGAN2, the normally distributed latent vector $z$ is passed through a dense network to create a different latent vector $w$ which was shown to have better properties in representation of higher order attributes, and this vector $w$ is then used by the ixmage synthesis model. 

%We made two important modifications to the architecture to utilize it for explaining a classifier: \gal{As above, find a different wording than 'modifications' in order not to create a feeling of some tweaks to something the reader already knows}
%\oran{I disagree, these are modifications for GAN training procedure, I think it's ok to assume the reader is familiar with it}.
%\begin{itemize}
Finally, we add an encoder which is trained to predict the latent vector $w$ from the image (see \figref{fig:architecture}). This is necessary for two reasons. First, it allows us to explain the classifier output on any given input image (and not only on GAN-generated images). Second, it allows us to ensure that the generative model captures classifier-related attributes by using the Classifier-Loss (see below).
    %The encoder is trained jointly with the generative model as described in \ref{subsection:training_expgan}.
    % \item The generative model is conditioned on the classifier output. The classifier output $c$ is concatenated to the $w$ latent vector before passing it into the generative model (See Fig. \ref{fig:architecture}). Adding this condition helps the style space to contain more attributes which are affect the classifier decision, as described in \ref{subsection:training_with_classifier}.

%\end{itemize}

\subsection{Training {\ours}}

\label{subsection:training_expgan}
\figref{fig:architecture} shows the components of {\ours} and its training procedure. The training method of our generative model is based on the standard GAN training procedure, but adds to it several modifications required for guiding the StyleSpace to contain classifier-related attributes. The basic GAN training recipe is to train the generator $G$ and an adversarial discriminator $D$ simultaneously~\cite{GoodfellowGAN}.
%\footnote{We also follow the procedure described in \cite{karras2020analyzing} by using the path regularization step every few training iterations.} 
Additionally, we jointly train an encoder $E$  with the generator $G$, using a reconstruction loss (i.e., the Encoder and Generator function together as an autoencoder). Finally, we add two components that introduce the classifier into the training procedure.
%This is done by adding a cross entropy loss between the input image used to generate the image and the output of the classifier on the generated image. 
%Additionally, as described above, we use the classifier output as an additional input for $G$.

%\vspace*{0.1cm}
\noindent
\inbarm{Not sure about the explanation I add for conditional training. Please take a look. Another option is to merge the motivation for both conditional training and classifier loss.}
{\bf Conditional Training:} We provide the generator with the intended value of the classifier output on the generated image. Adding this condition helps the StyleSpace to contain more attributes that effect the classifier's decision (as the StyleSpace coordinates become an affine transformation of the classifier output). %This conditional mode of training allows the StyleSpace to encode classifier-related attributes that are distinct from the class label itself, since the latter is provided as input.
%\oran{I don't agree with the last sentence. Suggested edit: We provide the generator with the intended value of the classifier output on the generated image. In case of the auto-encoder setting, this conditional input is the output of the classifier on the original image. This allows the StyleSpace to directly encode classifier related attributes, as it now becomes an affine transformation of the classifier output.}

\noindent
{\bf Classifier Loss:}
A GAN trained on a set of images will not necessarily capture visual structures related to a particular classifier. For example, a GAN trained on retinal images will not necessarily visualize pathologies corresponding to a particular disease. In this case, it will be impossible to visually explain a classifier for this pathology using this GAN. To overcome this difficulty, we add a Classifier-Loss on the images generated by the GAN, during the GAN training. This loss is the KL-divergence between the classifier output on the generated image, and the classifier output on the original input image. This loss ensures that the generator does not ignore important details which are meaningful for the classification, or collapse into only one of the labels.

The overall {\ours} training loss is the sum of the losses:
\begin{equation}
\label{eq:all_loss}
    Loss = \mathcal{L}_{adv} + \mathcal{L}_{reg} + \mathcal{L}_{rec} + \mathcal{L}_{cls}~,
\end{equation}
where $\mathcal{L}_{adv}$ is the logistic adversarial loss \cite{GoodfellowGAN}, and $\mathcal{L}_{reg}$ is the path regularization described in \cite{karras2020analyzing}. The encoding reconstruction loss, 
$\mathcal{L}_{rec}$, is given by $\mathcal{L}_{rec}^{x}+ \mathcal{L}_{LPIPS}+ \mathcal{L}_{rec}^w$
% \begin{equation}
% \mathcal{L}_{rec} = \mathcal{L}_{rec}^{x}+ \mathcal{L}_{LPIPS}+ \mathcal{L}_{rec}^w,
% \end{equation}
where the first two terms are calculated between the input image $x$ and the conditioned reconstructed image $x' = G[E(x), C(x)]$. More specifically, $\mathcal{L}_{rec}^{x} = \lVert x' - x \rVert_1$ and $\mathcal{L}_{LPIPS}$ is the LPIPS distance between $x$ and $x'$ as described in \cite{zhang2018perceptual}. The third term,  $\mathcal{L}_{rec}^{w}$, is adapted from the style reconstruction loss in \cite{choi2020stargan}: $ \mathcal{L}_{rec}^{w} = \lVert E(x') - E(x) \rVert_1$.
Finally, the Classifier-Loss is  
% \begin{equation}
% \label{eq:loss_cls}
     $\mathcal{L}_{cls} = D_{KL} \left[ C(x')|| C(x) \right]$.
% \end{equation}
% where $C$ is the classifier model which returns a vector of logits.
In \secref{sec:eval_and_result} we provide ablation results on these losses.

\subsection{Extracting Classifier-Specific Attributes}
\label{sub:discovering_attributes}

Thus far we trained a generative model that is constrained to capture classifier-related information. We next turn to finding coordinates in the StyleSpace of the model, which encode classifier-specific attributes. Namely, we seek specific coordinates in the StyleSpace such that changing them will change the generated image in a way that alters the classifier output in a non-negligible way. This will enable generating counterfactual explanations for a given image.

\newcommand{\attproc}{\mbox{\emph{AttFind}}}
Algorithm~\ref{alg:attfind} describes the {\attproc} procedure for discovering classifier-specific attributes. Denote by $K$ the dimension of the style vector (across all layers), and by $C(x)$ the vector of classifier logits (pre-softmax probabilities) for image~$x$.
{\attproc} takes as input the trained model and a set of $N$ images whose predicted label by $C$ is different from $y$. For each class $y$ (e.g., $y$=``cat'' or $y$=``dog''), {\attproc} then finds a set $S_y$ of $M$ style coordinates (i.e., $S_y\subset [1,\ldots,K]$ and $|S_y|=M$), such that changing these coordinates increases the average probability of the class $y$ on these images.\footnote{This may be viewed as an estimate of the Average Causal Effect \cite{pearl2009causality}.} Additionally it finds a set of ``directions'' $D_y\in\{\pm 1\}^M$ that indicate in which direction these  coordinates need to be changed to increase the probability of $y$. \amirg{increase or decrease?...} \amirg{say something about change magnitude? Is it different for train or test?} 

\begin{figure}[th]
%\vspace{-.05in}
   \centering
   \includegraphics[width=\linewidth]{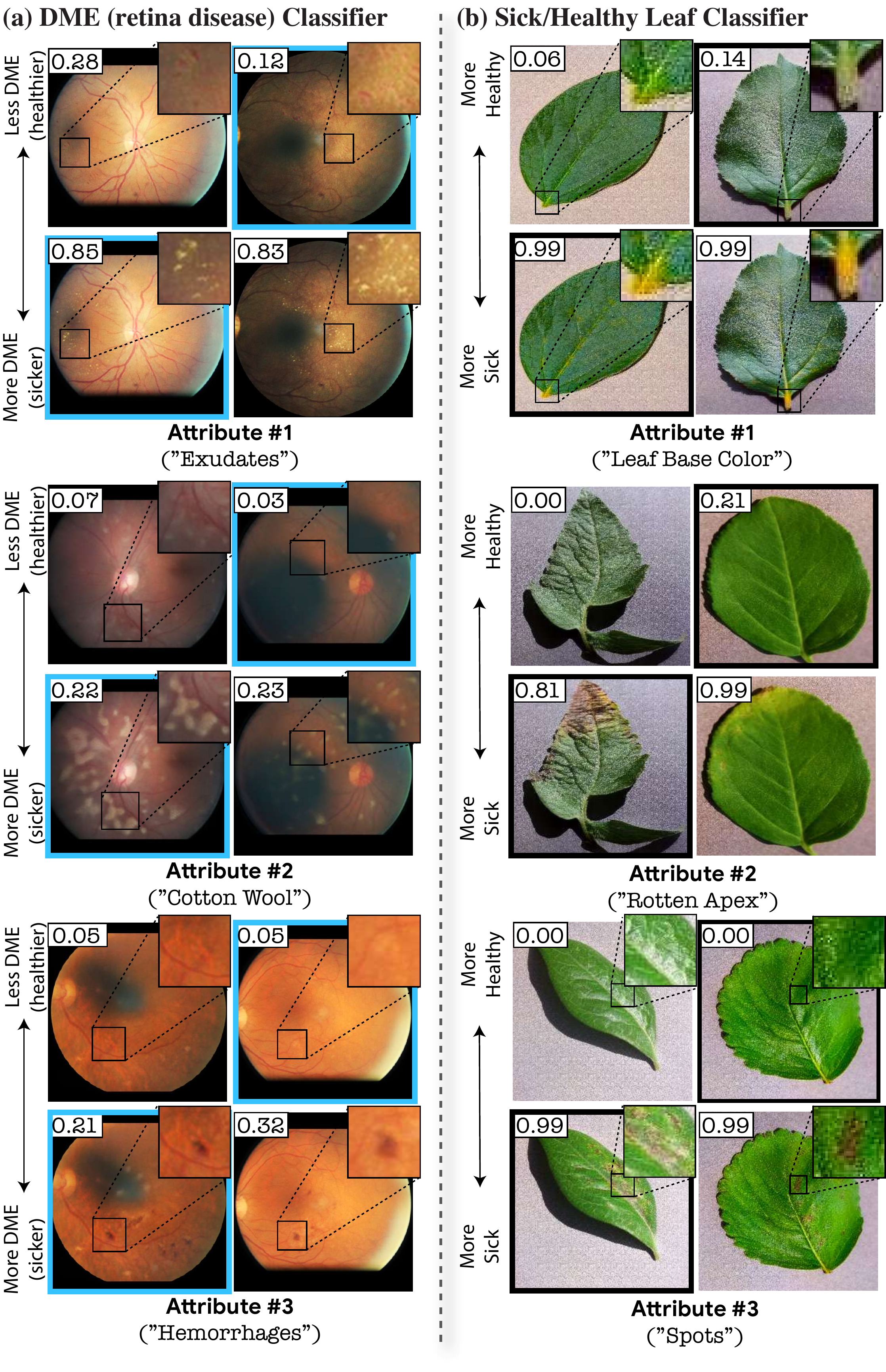}
   \vspace{-.25in}
   \caption{{\bf Top-3 automatically detected attributes for (a) DME classifier of retina images and (b) Classifier of Sick/Healthy leaf images.} The respective classifier scores are presented in their top left corner. The generated counterfactual examples are marked by a frame. The top discovered attributes for both classifiers turn out to be well aligned with known disease indicators ( \cite{rakhlin2018diabetic}, \cite{hughes2015open}). Please see animated-GIFs in the \href{http://explaining-in-style.github.io/supmat.html}{project website results page} to view these counterfactual changes (explanations) dynamically.}
   \label{fig:retina_top_5_attributes}
\vspace{-.11in}
\end{figure}

\begin{figure}[t]
%\vspace{-.05in}
   \centering
   \includegraphics[width=0.98\columnwidth,scale=0.9]{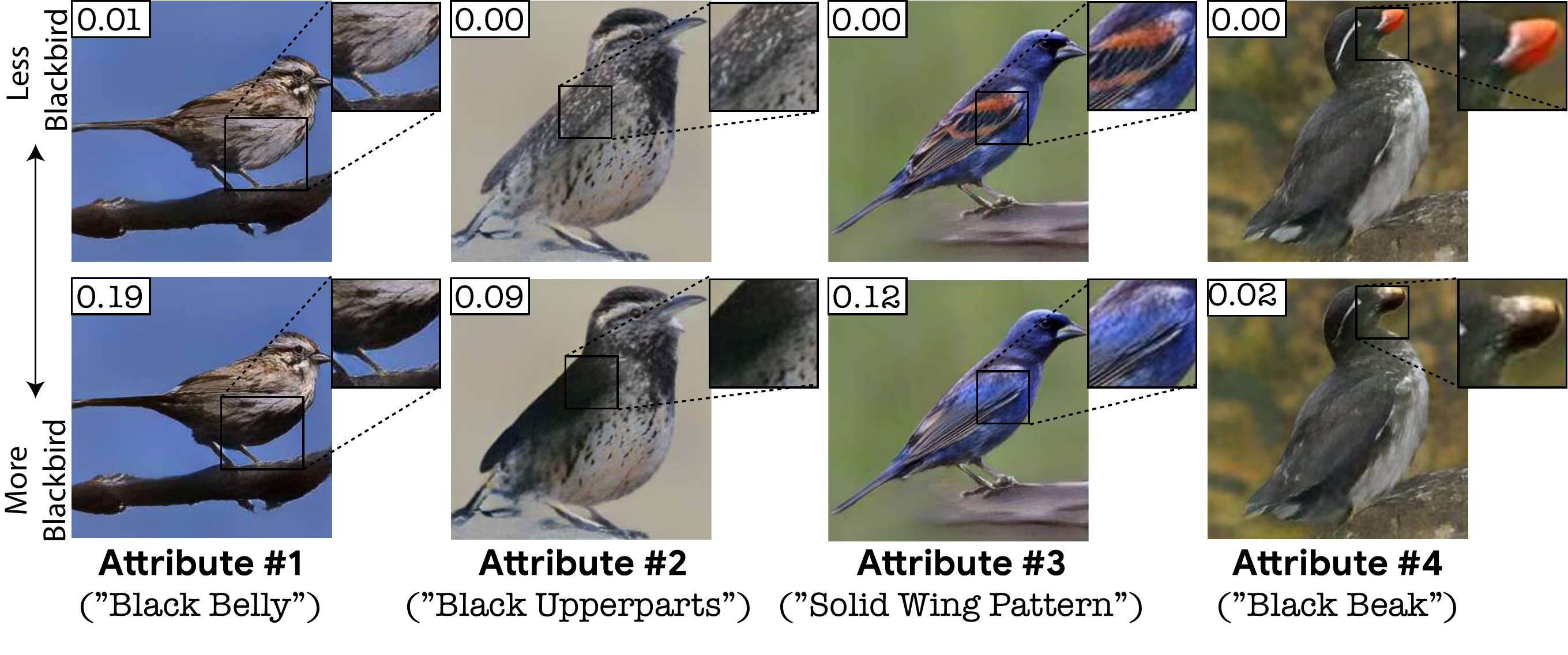}
   \caption{{\bf Explaining multi-class classifiers: top-4 automatically detected attributes for the class {\em brewer blackbird} in a 200-way classifier trained on CUB-2011 \cite{wah2011caltech}.} The classifier scores for the {\em brewer blackbird} class are presented in the top left corner. The top discovered attributes correspond to attributes in the CUB taxonomy. Please see \href{http://explaining-in-style.github.io/supmat.html}{project website results page} for additional results as well as animated-GIFs.}% to view these counterfactual changes (explanations) dynamically.}
   \label{fig:cub_attributes}
\vspace{-.1in}
\end{figure}

%{\attproc} discovers attributes one at a time, discarding images explained by found attributes to avoid duplicates, and maximize explanation coverage. Thus, 
{\attproc} proceeds as follows: At each iteration it considers all $K$ style coordinates and calculates their effect on the probability of $y$.\footnote{More precisely, we use logits instead of probabilities, as often preferred in classifier explanations, e.g. \cite{shrikumar2017learning, dhamdhere2018important}.} \footnote{If a coordinate has an inconsistent change direction it is discarded. \label{footnote_ambiguous}} It then selects the coordinate with largest effect, and removes all images where changing this coordinate had a large effect on their probability to belong to class $y$ 
(i.e., this coordinate suffices to ``explain'' those images; no need to proceed to other coordinates).
%\michali{Need to discuss this last sentence}. 
%(e.g., this coordinate alone suffices to flip the classification of these images).
This is repeated until no images are left, or until $M$ attributes are found. The process is summarized in Algorithm~\ref{alg:attfind}.
Examples of these automatically detected attributes, for a variety of different classifiers (binary and multi-class), are found in Figs.~\ref{fig:gender_top5_attriubtes},\ref{fig:retina_top_5_attributes}, \ref{fig:cub_attributes}.
%, .\amirg{specify if this is class dependent} 
\vspace{-.1in}
\begin{algorithm}
 \caption{AttFind \label{alg:attfind}}
\SetAlgoLined

\SetKwInOut{Init}{Initialization}{}{}
\KwResult{
{Set $S_y$ of top $M$ style coordinates  \&  set $D_y$} of their directions.}
\KwData{{Classifier $C$. A set $X$ of images whose predicted label  by C is \underline{not} $y$.
 Generative model $G$. Threshold $t$.}}
 \Init{$S_y$, $D_y$ = empty.}
 \While{$|S_y| < M$ or $|X| > 0$}{
  \For{$x$ in $X$}{
   \For{style coordinate $s \notin S_y$}{
     Set $\tilde{x}$ to be the image $x$ after changing coordinate $s$ in directions $d\in\{\pm 1\}$\;
     Set $\Delta[x,s,d] = C_y(\tilde{x}) - C_y(x)$\;
    %  compute the difference of the classifier prediction for changing style index $s$ in both directions $d\in\{\pm 1\}$ for image $I$\;
    }
  }
%  Set $\Delta{M}[s,d]=reduce\_mean(R,axis=0)$. \;
Set $\bar{\Delta}[s,d] = Mean(\Delta[x,s,d])$ over all $x\in X$\;
%   compute the average effect on all images $I_y$ for every style index $s$ and the two directions $d$.\;
  \For{style coordinate $s \notin S_y$}{
    \uIf{$\bar{\Delta}[s,1]>0$ \&  $\bar{\Delta}[s,-1]>0$ $^{\ref{footnote_ambiguous}}$}{
%  {\mbox{Set both to Zero: \   $\bar{\Delta}[s,1]=0$, $\bar{\Delta}[s,-1]=0$\;}}{}
    \mbox{set to Zero:   $\bar{\Delta}[s,1]=0 \  \& \  \bar{\Delta}[s,-1]=0$\;}}{}
  }
  Set $s_{max},d_{max}$ to be $\arg\max_{s,d} \bar{\Delta}[s,d]$\;
  Add $s_{max}$ to $S_y$, and $d_{max}$ to $D_y$\;
  %Set $S_y=S_y\cup \{s_{max}\}$ and $D_y=D_y\cup \{d_{max}\}$\;
  Let $X_{explained}$ be all $x\in X$ s.t. ${\Delta}[x, s_{max},d_{max}]>t$\;
  Set $X = X\setminus X_{explained} $\;
}
\end{algorithm}
%\vspace*{-0.5cm}
\comment{
\begin{algorithm}
\SetAlgoLined
\SetKwInOut{Init}{Initialization}{}{}
\KwResult{\mbox{Top $M$ style coordinates $S_y$ \& directions $D_y$}}
\KwData{\mbox{Classifier $C$. A set of $N$ images 
 $X_y$ predicted as}  \mbox{class $y$ by $C$ .  
 Generative model $G$. Threshold $t$.}}
 \Init{$S_y$, $D_y$ = empty.}
 \While{$|S_y| < M$ or $|X_y| > 0$ }{
  \For{$x$ in $X_y$}{
   \For{style coordinate $s \notin S_y$}{
     Set $\tilde{x}$ to be the image $x$ after changing coordinate $s$ in direction $d\in\{\pm 1\}$\;
     Set $\Delta[x,s,d] = C_y(x) - C_y(\tilde{x})$\;
    %  compute the difference of the classifier prediction for changing style index $s$ in both directions $d\in\{\pm 1\}$ for image $I$\;
    }
  }
%  Set $\Delta{M}[s,d]=reduce\_mean(R,axis=0)$. \;
Set $\bar{\Delta}[s,d]$ =  Mean $(\Delta[x,s,d])$ over all $x\in X_y$\;
%   compute the average effect on all images $I_y$ for every style index $s$ and the two directions $d$.\;
  \For{style coordinate $s\notin S_y$}{
  \uIf{$\bar{\Delta}[s,1]>0$ and $\bar{\Delta}[s,-1]>0$}{
   Set  $\bar{\Delta}[s,\pm 1]=0$\;
  }{}
  }
  Set $s_{max},d_{max}$ to be $\arg\max_{s,d} \bar{\Delta}[s,d]$\;
  Add $s_{max}$ to $S_y$, and $d_{max}$ to $D_y$\;
  %Set $S_y=S_y\cup \{s_{max}\}$ and $D_y=D_y\cup \{d_{max}\}$\;
  Let $X_{explained}$ be all $x\in X_y$ s.t. ${\Delta}[x, s_{max},d_{max}]>t$\;
  Set $X_y = X_y\setminus X_{explained} $\;
%  Remove from $I_y$ images that this style coordinate had a significant effect on the probability of $y$.\;
 }
 \caption{AttFind \label{alg:attfind}}
\end{algorithm}
}
\vspace{-.2in}
\comment{
To find $S_y,D_y$ {\attproc} calculates the average change in probability corresponding to each coordinate of the StyleSpace. 
 It receives the trained model and a set of images, and seeks a set of style coordinates whose modification directly affects classifier output. In order to find a robust and small set, our search procedure includes the following components: a) Any image ``explained'' by an attribute discovered is removed from the training set, so as to avoid finding copies of the same attribute. b) Decreasing and increasing coordinate value results in 

We find attributes using the StyleSpace \cite{wu2020stylespace} (abbreviated as S-space), the space of channel-wise style parameters. This space is a linear transformation of $w$, with much larger size, which is used as the style vector for each layer in the network. 
As was shown in \cite{wu2020stylespace}, moving in each dimension of the S-space results in disentangled GAN attributes. Here we exploit this phenomenon to detect directions which affect the classifier decision. Since the training of our {\ours} generator is constrained by the classifier loss, we harness this inherent property to extract disentangled classifier-specific attributes which visually explain the classifier decision.

To find such coordinates, we take a large number of images, N (in our experiments we typically used $N=1000$), with uniform distribution over predicted classes. For each of those N images we shift each element in the S-space to its maximum and minimum value, and measure the induced change in the classifier prediction. To avoid generating images with artifacts when moving elements in the S-space, we use the discriminator that was trained with the generator to sift out those images. We ignore the change in the classifier predictions for those images that (according to the discriminator) moved too far from the generator's distribution.

Next, for each class we seek the top-$M$ most dominant style directions (attributes) that dominate its classification (where $M$ is a predetermined number). We split the N images into classes based on the classifier prediction.\amirg{but they were already split to begin with. Because we say we select them such that they are 50/50.} For each class we retrieve the style directions that significantly decrease the probability of being in that class.
%(and increase the probability of belonging to one of the other classes).
Retrieving the style coordinates and signs is done as follows. We compute the mean prediction change for all class images, and choose the style direction with the largest mean change. Once this style index and its movement direction are found, we verify that when moving in the opposite direction in this style index, the mean change is not positive. We remove images for which this style direction results in a significant change in logits. We repeat this process until we reach the predefined number M of style directions, or when all class images were covered (had a significant change in the classifier predictions by one of the style directions).
These retrieved style directions form the basis for for our \emph{classifier-based counter-factual explanations}. Examples of the top-M automatically detected attributes, for a variety of different classifiers, can be found in Figures ~\ref{fig:gender_top5_attriubtes}, \ref{fig:retina_top_5_attributes}.
}

\inbarm{Note that {\attproc} can be applied to multi-class }

\comment{
\subsection{Using {\ours}  to Visually Explain the Classifier}
In this section we first explain how we extract classifier-specific attributes from the StyleSpace\amirg{change to style-space throughout?} of our {\ours} GAN ( \secref{sub:visualizing_classifier}). These classifier-specific attributes can then be used to produce visual explanations of the classifier decisions via \emph{counter-factual examples}.

In \secref{sub:visualizing_image} we further show that not only can we extract typical \emph{classifier-specific attributes}, but can further extract also \emph{image-specific attributes}. Namely, highlight attributes which explain why the classifier made a specific prediction for a specific image.
}
%Fig. \ref{fig:single_image_attributes}
\subsection{Generating Image-Specific Explanations}
\label{sub:visualizing_image}
 {\ours} provides a natural mechanism for explaining the decision of a classifier on a specific image: simply find {\ours} attributes that affect the classifier's decision on this image, and visualize the effect of changing those.

%If we are interested in explaining how to flip the classifier's decision on this image (e.g., in the binary case, what made the softmax output cross the $0.5$ threshold), we can further seek a minimal subset of attributes whose joint change will flip the decision.

There are various strategies for finding a set of image-specific attributes. The simplest is to iterate over {\ours} attributes, calculate the effect of changing each on the classifier output for this image, and return the top-k of these. We can then visualize the resulting $k$ modified images. We refer to this strategy as {\bf Independent} selection. Alternatively, it could be that individual attributes do not have a large effect, and thus we can search for a set of $k$ {\ours} attributes, whose {\em joint} modification maximizes classifier change. In order to avoid checking all possible $O(2^k)$  subsets, we perform a greedy search (i.e., at each step find the next most influential attribute for this image, given the subset of attributes selected so far; halt once the classifier has flipped its classification). We can then visualize the effect of modifying this subset. We refer to this as {\bf Subset} selection. 
%See Fig.~\ref{fig:single_image_attributes} and \ref{fig:single_image_leaf}, for examples of these approaches.
\comment{
In principle, it might be that no such attributes exist for explaining a specific image. 
However, the {\attproc} procedure is constructed to find attributes that affect many images. \amirg{check this} 
Our experiments show that in practice the subset of attributes our algorithm finds suffices to explain classifier decisions for most images (\tabref{table:sufficiency}). \yossig{here should be a change in terminology to "effects the classifier".} 
Such a sufficiency criterion is measured and explained in \secref{sub:effect_on_classifier} and \tabref{table:ablation_sufficiency}.
}
%\comment{

\begin{figure*}[th]
   \centering
   \vspace{-.1in}
   \includegraphics[width=0.9\linewidth]{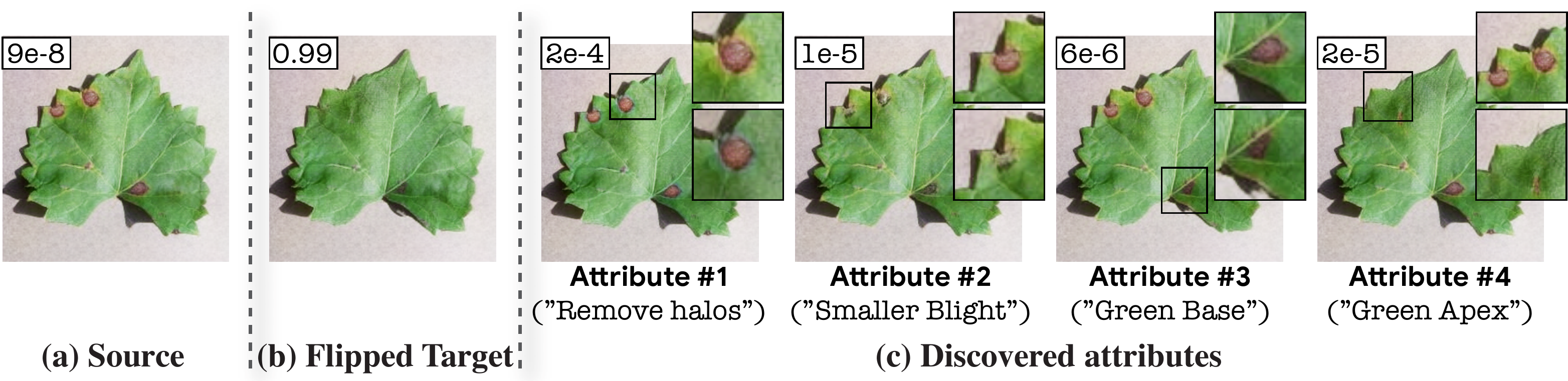}
%   \vspace{-.2in}
   \caption{{\bf Combining several attributes}.
%   These attributes were selected by the {\bf Subset} selection method described in \secref{sub:visualizing_image}, so that their combination flips the classifier decision.
   Attributes in {\bf (c)} are selected by the {\bf Subset} method to inflict the largest accumulated effect on the classification of the image in {\bf (a)}. Interventions on individual attributes result in a small change of the classifier output, yet intervening on all of them results in image {\bf (b)} where the classification is flipped. The classifier probabilities of healthy are presented in the top-left corner.}
   \label{fig:single_image_leaf}
   \vspace{-.15in}
\end{figure*}

Figures \ref{fig:single_image_attributes} and \ref{fig:single_image_leaf} show examples of image-specific explanations (one per selection strategy; see \secref{sec:eval_and_result}). We ask what has led the classifier to classify this person as ``Perceived Old'' and not ``Perceived Young'', or this leaf as ``Sick'' and not ``Healthy''. The figures show the top \emph{image-specific} attributes that drove the classifier to its prediction on this image.
%}
%classify this image as it did.
%In \secref{} we provide examples of this procedure.\amirg{check we do}
%enables us to answer the question - what is the minimal number of attributes we should change in this image so that the classifier decision would change. This is done by generating counterfactual images which change very few attributes in the original image, yet result in a flip the classifier decision.
\comment{
Fig. TODO shows such an example. We can ask what made the classifier decide to classify a specific dog image as "Dog" (as opposed to "Cat"). The figure shows the attributes that the classifier believes are most prominently "doggish": 
\begin{itemize}
    \item TODO add according to the image
\end{itemize}
We further show that it is enough to change a small number of attributes in order to flip the classifier prediction on a specific image from class A to class B. \figref{fig:single_image_leaf}
Finally, we show that different combinations of subsets of attributes (different "paths"), can be used in order to flip the classifier prediction on a specific image from class A to class B. We refer to this as a "diversity" of possible explanations. Such an examples is shown in \figref{fig:diversity}
}
%\amirg{remember to refer to counterfactual visual explanations and to the work from Google that explains with attributes, and also to GANalyzer}
% \subsection{Implementation Details}

% \input{fig_single_image}
% \input{fig_single_image_leaf}
% \section{Sufficiency \& Diversity of the Attributes}
\begin{table}[th]
%\vspace{0.2cm}
    \centering
    \scalebox{0.9}{
    \begin{tabular}{c|c}
\toprule%\hline
    {\bf Dataset} & {\bf Classifier} \\
\midrule%\hline
     AFHQ \cite{choi2020stargan} & Cats / dogs \\
%\hline
     AFHQ & Wild cats species \\
%\hline
     FFHQ \cite{karras2019style} & Perceived gender \\
%\hline
     FFHQ & Perceived age \\
%\hline
     Plant-Village \cite{hughes2015open} & Healthy / sick leaves \\
%\hline
     Retinal Fundus \cite{krause2018grader} & DME / non-DME \\
%\hline
     CUB-2011 \cite{wah2011caltech} & Bird species \\
% \hlinehttps://www.overleaf.com/project/601aaac2a2319087bce0e44c
% LSUN \cite{yu2015lsun} & Kitchen / Bedrooms \\
\bottomrule%\hline
    \end{tabular}}
    \caption{List of datasets used in this paper.}
    \label{tab:domains}
    \vspace*{-0.3cm}
\end{table}

\section{Evaluation and Results}
\label{sec:eval_and_result}
We test our {\ours} method on a variety of classifiers from a diverse set of domains, as listed in \tabref{tab:domains}. The classifiers are based on the MobileNet \cite{Mobilenet} architecture, and achieve a high accuracy of at least 95\% on their test sets. {In the results below, we show images corresponding to changing visual attributes found by \ours{}. It is important to emphasize that we do not imply these modified images necessarily correspond to a modified label probability in reality, but rather that they result in modification of classifier output. In other words, the modified classifier output may reflect biases in classifier training, and not a true correlation between the label and visual attributes.}

\subsection{Qualitative Evaluation}
We first demonstrate that each {\ours} attribute corresponds to clear visual concepts, and then show that these can be used to explain classifier outputs on specific images.
\paragraph{Visualizing {\ours} Attributes.}
We begin by showing that the attributes discovered by {\ours} indeed correspond to coherent semantic notions (see \secref{sec:user_study} for user studies that further demonstrate this). We emphasize that we do not choose attributes by manual inspection but rather use the top attributes found by the {\attproc} procedure. 

The semantic meaning of the attribute already becomes clear even when inspecting its effect on a pair of images: one where its effect is increased, and one where it's decreased. Figures \ref{fig:gender_top5_attriubtes} and \ref{fig:retina_top_5_attributes} demonstrate this for four domains. For each domain we consider the top {\ours} attributes. For each attribute, we find an image in each class where attribute modification was significant, and display the original and modified image. 
% See supplementary for full results and animated-GIFs, which provide much clearer \emph{dynamic} counterfactual explanations.
See results and animated-GIFs in \href{http://explaining-in-style.github.io/supmat.html}{project website results page} to view the these counter-factual explanations \emph{dynamically}.

\begin{figure*}[th]
   \centering
   \includegraphics[width=0.9\linewidth]{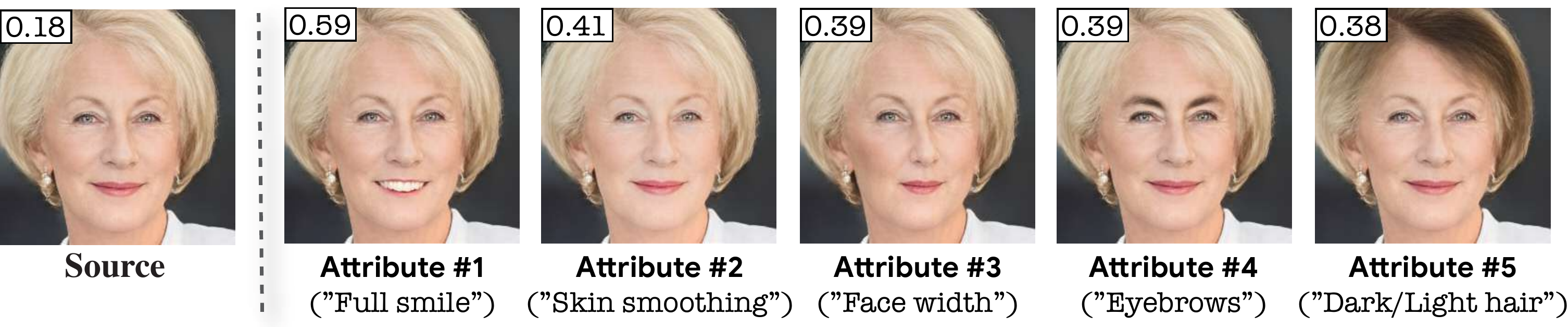}
%   \vspace{-.2in}
   \caption{{\bf Image-specific explanations:} Top-5 automatically detected attributes for explaining a perceived-age classifier for a specific image using the {\bf Independent} selection strategy. Attributes are sorted by their effect on the classification of the specific image, resulting in different attributes from those presented in \figref{fig:gender_top5_attriubtes} which have the largest average effect over the entire dataset. The classifier probabilities of young are shown in the top-left corner.}
   \label{fig:single_image_attributes}
\end{figure*}

It can be seen that each of the top attributes extracted by {\ours} is visually-interpretable. Additionally, modifying each attribute leads to a significant change in the classifier output. The quality of the explanations that {\ours} provides can be demonstrated in the cases of the healthy/sick leaves classifier and the retina DME classifier (\figref{fig:retina_top_5_attributes}) where the top discovered attributes are aligned with disease indicators that appear in the corresponding datasets (\cite{hughes2015open, rakhlin2018diabetic}).
%We first demonstrate the top attributes found by \ours{} for the classifiers in \tabref{tab:domains}.
\comment{
{\bf Classifier-Related Attributes}
\oran{A major benefit of \ours{} is that can generate counterfactual explanation for a single image, which provides a powerful tool for ``debugging'' the classifier decision on it, as described in \secref{sub:visualizing_image}. We describe two variants of this method, each providing different observation.
First, we can find the top attributes on one specific image. Essentially this means what does the classifier looks at on this image, which might be different than the top attributes affecting the classifier decision on average. \figref{fig:single_image_attributes} (a) shows an example where the top $5$ attributes affecting this specific image are not the same attributes found by {\attproc} on the Age classifier as shown in \figref{fig:gender_top5_attriubtes}.
Second, we can use {\ours} to break down the decision of the classifier into different components.  \figref{fig:single_image_attributes} (b) demonstrates this on a case of a sick leaf. }
}
\comment{
\bf{Visualizing Classifier-Related Attributes}
%\label{sub:visualizing_classifier}
After training the generative model and discovering the StyleSpace coordinates (Sections \ref{subsection:training_expgan} and \ref{sub:discovering_attributes}), we take specific images and change these coordinates to change the classifier decisions. Figures \ref{fig:retina_top_5_attributes} and \ref{fig:gender_top5_attriubtes} show the results for four different domains. It can be seen that each attribute is visually-interpretable and changing it leads to a change in classifier output. For example, the top few attributes that were discovered for the healthy/sick leaves classifier are aligned with a set of different plants diseases that appear in the dataset \cite{hughes2015open}.}
\comment{
 \ref{fig:gender_top5_attriubtes} show attributes from classifiers we trained on multiple domains, which were extracted by the method described in subsection \ref{sub:discovering_attributes}. These attributes serve as an explanation for these models, in the sense that each attribute demonstrates one factor which affect the classifier decision. For example, the top few attributes that were discovered for the healthy/sick leafs classifier, are aligned with a set of different plants diseases that appear in the dataset \cite{hughes2015open}.}
% One can say that in order to change the decision of the classifier you can move the image along one or more of these axes. \yossig{This already appeared before.}
% Note that our method does not claim to be a full explanation for the classifier, in the sense we do not claim that knowing these attributes is enough to predict the classifier decision on any image. 
% TODO add a completeness measure.
\paragraph{Explaining multi-class classifiers} \attproc{} is also applicable to multi-class problems. \figref{fig:cub_attributes} demonstrates this on a classifier trained on CUB-2011 (200 classes) \cite{wah2011caltech}. Indeed, we observe that \ours{} detects attributes that correspond to attributes in CUB taxonomy.

\paragraph{Providing Image-Specific Explanations.}
Thus far we showed that each {\ours} attribute corresponds to an interpretable visual concept. We can now use these to provide counterfactual explanations for specific images. Namely, for a given image we can provide statements such as: ``had you changed attribute $\#1$ and $\#3$, the classifier output would have changed''. Since attribute $\#1$ and $\#3$ have clear semantics (e.g., $\#1$ is adding a moustache) the counterfacutal explanation is easy to understand and is informative.  
% \phillip{general comment: I think we should consider both classifier-level explanations and image-level explanations to be about ``counterfactuals" since the intro is written as if our whole framework is about counterfactual explanations, rather than just the image-level explanations. Although I agree that image-level explanations are the most obvious kind of counterfactual...}
%In addition to finding attributes which affect the classifier in general, we can also use the \ours{} framework for explaining the decision on a single image. 
To find a counterfactual explanation, we use the method in \secref{sub:visualizing_image}. % to find a small set of attributes whose change will result in changing the decision of the classifier. 

\figref{fig:single_image_attributes} illustrates this for the ``Perceived Age'' classifier, where we use the {\bf Independent} selection method. It can be seen that there are five attributes that individually affect that classifier output considerably. Also, these are not the attributes with largest average effect across many images (see \figref{fig:gender_top5_attriubtes}), but rather those that most affect this {\em specific} image.
\figref{fig:single_image_leaf} shows an example for the Plants domain, where we use the {\bf Subset} selection method. Here each of the selected attributes has a smaller effect (though the change in logit is significant), but the combined effect of changing the three attributes results in flipping the classifier decision. These examples nicely demonstrate that {\ours} can be used to ``decompose''  classifier decisions into a set of visual attributes. 
%As shown in \figref{} \amirg{complete}
%which demonstrate what changes on this image also change the classifier outcome.
%\figref{fig:single_image_attributes} shows examples for multiple changes in a single image, explaining what is the set of manipulations for a specific image that can cause the classifier to change its decision on it.
% \subsection{Discovering biases in the dataset}
% Another important use case of our method is discovering biases in the classifier, meaning that our classifier is affected by attributes which it should be indifferent to. For example, our data is biased such that one class contains more images with a certain unrelated feature, that bias could be picked up by a classifier. Our method is able to detect such a bias and present it to the designer of the classifier.
% TODO To demonstrate this, we made an experiment where we added a small watermark to $1\%$ of the images in one class of the healthy / sick leaves dataset, and trained a classifier on this data. As shown by figure \ref{fig:single_image_attributes} this attribute is detected as part of our method.
% \input{tab_sufficiency.tex}

\subsection{Quantitative Evaluation}
%\phillip{repetitive wording (same wording as the last section):} Thus far we provided qualitative results visualizing \ours{} attributes.
%We next turn to quantiti evaluation of {\ours} attributes. 
It is not immediately clear how to evaluate multi-attribute counterfactual explanations. However, the three following criteria seem key to any such method:
%We consider the following key criteria: 
%In our explanations, attributes are demonstrated via interventions (i.e. changing the value of a StyleSpace coordinate) and observing the modified images. Hence to evaluate the quality of attributes extracted by \ours{}, we define three desired properties based on such interventions: 
\\
\textbf{Visual Coherence.} Attributes should be identifiable by humans. For instance, the effect of a coordinate that controls pupil dilation in cats can be easily understood by humans after seeing a few examples. On the other hand, if the coordinate changes different visual attributes for each image (e.g. dilates pupils in some images, while shortening ears in others) then understanding its effect is a more difficult task, resulting in a less coherent visual attribute. 
\\
\textbf{Distinctness.} Extracted attributes should be distinct. Having distinct attributes lets us compose several counterfactual explanations that expose different elements underlying classifier decisions (e.g., as opposed to GANalyze \cite{goetschalckx2019ganalyze}).
%On the other hand attributes that are very similar provide explanations that are not as rich\yossig{Suggesting: and collapse to the class, as shown in figure \ref{fig:motivation} for the GANalyze method \cite{goetschalckx2019ganalyze}}. 
%For instance if all the extracted attributes control the snouts in dog and cat images, they might have a large effect on classification, yet they only correspond to a single.
%\yossig{Maybe we should say here again something about GANalyze. Also - I think that the example should be more extreme -  "For instance if all the $s$ coordinates change fully the images from dog to cat, they don't explain...}\phillip{agreed, except do we have experiments comparing to GANalyze?}
\\
\textbf{Effect of Attributes on Classifier Output.} Changing the value of attributes in an image should result in a change in classifier output. Furthermore, different attributes should have complementary effects so that modifying multiple attributes will result in flipping the decision of the classifier on most images.


\subsubsection{Baselines and Model Variants}
\label{sub:baselines}
As discussed in \secref{sec:related}, to-date, multi-attribute counterfactual explanations of visual classifiers have not yet been achieved for real images. Thus, there are no directly comparable baselines in the literature. However, most closely related to our method is the original StyleSpace defined in \cite{wu2020stylespace}. While \cite{wu2020stylespace} was not proposed as a method to explain a classifier, we can use it as a baseline to test  two key components our method adds to the StyleSpace framework: (i) classifier-specific training (CST) of the StyleGAN and (ii) the \textit{AttFind} method for finding classifier-related coordinates in StyleSpace. To test the importance of these two contributions, we compare against \ours~ without CST and also against using the StyleSpace selection method in \cite{wu2020stylespace}:%, which uses neither CST nor \textit{AttFind}:
\begin{itemize}[topsep=0pt,itemsep=-1ex,partopsep=1ex,parsep=1ex]
\vspace*{0.2cm}
\item \textbf{{\ours} w/o Classifier-Specific Training (CST):} The training procedure for {\ours} incorporates the classifier into the StyleGAN training procedure to obtain a classifier-specific StyleSpace.  Here we consider the effect of using our \attproc{} procedure with a standard StyleGAN2 that does not involve the classifier. 
\vspace*{0.2cm}
\item \textbf{{\sspacename} \cite{wu2020stylespace}}: %While \cite{wu2020stylespace} was not proposed as a method to explain a classifier, we nonetheless test if it can be applied to this purpose.
\cite{wu2020stylespace} proposed identifying StyleSpace coordinates that relate to known visual labels. These coordinates can be identified by measuring the normalized difference between the coordinate values on each of the labels. As a baseline we use their method to find the top-M classifier-related coordinates in standard StyleGAN2 StyleSpace and compare against our method, which instead uses \textit{AttFind} and a StyleSpace trained with CST.
\end{itemize}

\subsubsection{A User Study for Coherence and Distinctness} \label{sec:user_study}
To evaluate coherence and distinctness we conducted a two-part user study. 
% Our evaluations include a user study examining the Visual Coherence and Distinctness in \ours{} and a quantitative evaluation of the attributes effect on the classifier, using a \emph{sufficiency} metric. \\
% \textbf{Evaluation of Sufficiency.} 
% \textbf{User Study for Coherence and Distinctness.} Our user study consists of two parts. 
% The first provides a quantitative evaluation of visual coherence and distinctness. 
The first part uses a setup similar to \cite{yeh2020completeness}.
% \footnote{Their study also asks users to identify attributes, though they do not use a generative model. Hence they show users patches of images to demonstrate attributes, instead of the animations we use here.}
Users are shown four animated GIFs, each corresponding to modifying an attribute for a given image. The left two GIFs are produced from an attribute $i$ and the right two GIFs are from attributes $i$ and $j$. The user is asked to identify the right GIF corresponding to attribute $i$  (see \href{http://explaining-in-style.github.io/supmat.html}{supplementary} for more information). We use the top $6$ style coordinates for \ours{} and for \sspacename, and perform the experiment using each of these sets.
%Each question shows the user two animations demonstrating an modification of a certain style coordinate applied to two different images. Then out of two candidate animations (one intervining on the same style coordinate that was demonstrated and the other on a different one), users are asked to select the candidate which changes the demonstrated attribute.
A correct answer shows that attributes are distinct, since users are able to distinguish between the extracted attributes. It also establishes visual coherence since users are able to classify animations as belonging to the attribute based only on two examples.
% Our user study further confirms the visual coherence and distinctness demonstrated in Fig.~\ref{fig:gender_top5_attriubtes}.
The results in \tabref{table:user_study} show that {\ours} achieves high accuracy in this task, suggesting attributes are distinct and coherent. It also outperforms {\sspacename} on all domains but Plants. However, on Plants we show in
\secref{sub:effect_on_classifier} that {\sspacename} attributes have little effect on the classifier.

% has results for the first part of the study, where users had to associate visualizations of the same attributes. 
\comment{
The high success rates prove that overall, users find visual changes caused by different style coordinates highly coherent and distinct. For the plants dataset, coordinates extracted by both methods correspond to attributes that are rather coherent and distinct, yet as discussed in \secref{sub:effect_on_classifier} the coordinates extracted by \ours{} have a much higher effect on the classifier.
}
\comment{
In tasks involving face images, we observe that the selection method of \cite{wu2020stylespace} finds $2-3$ 
coordinates which make a distinct and coherent change to the image (recognized by at least 90\% of the users)
% \footnote{i.e. they are recognized correctly by $90\%$ of the users or more}
, yet the other coordinates often make a negligible effect over both the image and classification, as implied by the high standard deviation reported in \tabref{table:user_study}. For the Cats/Dogs dataset we observe this behavior for all attributes extracted by \sspacename. The results emphasize the importance of \algref{alg:attfind} that chooses coordinates for which an \emph{intervention} affects the classifier; this is opposed to choosing coordinates whose \emph{values} differ between classes, but where an intervention might not have an effect. We provide examples of the results described above in the supplementary material.
}
% \begin{table}[ht]
%     \centering
%     \scalebox{0.7}{\begin{tabular}{|l|c|c|c|c|} \hline
%      &
%     Perceived Gender &
%     Perceived Age &
%     Plants & Cats \& Dogs\\
%     \hline
%     Ours & 0.96 ($\pm 0.047$) & 0.983 ($\pm 0.037$) & 0.916 ($\pm 0.068$) & 0.933 ($\pm 0.05$)\\
%     \hline
%     Wu \emph{et al.} \cite{wu2020stylespace}& 0.783($\pm 0.186$) & 0.85 ($\pm 0.095$) & 0.91($\pm 0.081$) & 0.65 ($\pm 0.18$)\\
%     \hline
%     \end{tabular}}
%     \vspace{0.3cm}
%     \caption{\textbf{User study results.} Fraction of correct answers on identification of the top-6 extracted attributes. Standard deviation is taken across scores of the different attributes.}
%     \label{table:user_study}
% \end{table}
\begin{table}[ht]
    \centering
    \scalebox{0.9}{\begin{tabular}{l|c|c} \toprule%\hline
     & \textbf{\sspacename} & \textbf{Ours}\\
    \midrule%\hline
    Perceived Gender & 0.783($\pm 0.186$) & 0.96 ($\pm 0.047$) \\
    %\hline
    Perceived Age & 0.85 ($\pm 0.095$) & 0.983 ($\pm 0.037$)\\
    %\hline
    Plants & 0.91($\pm 0.081$) & 0.916 ($\pm 0.068$)\\
    %\hline
    Cats/Dogs & 0.65 ($\pm 0.18$) & 0.933 ($\pm 0.05$)\\
    \bottomrule%\hline
    \end{tabular}}
    \vspace{0.3cm}
    \caption{\textbf{User study results.} Fraction of correct answers on identification of the top-6 extracted attributes.}
    \label{table:user_study}

\end{table}

%Standard deviation is taken across scores of the different attributes.\phillip{I confused as to what ``Standard deviation is taken across scores of the different attributes" means. Do we need to say this?}
% \yoavw{Once we have results for verbal test, add another figure/table. We'll have to decide exactly what to show there according to the answers we'll get.}
% We choose the top attributes to present.
For the second part of the study (performed on a different set of users) we show 4 GIFs demonstrating an intervention on a single style coordinate. Users are then asked to describe in 1-4 words the single most prominent attribute they see changing in the image. We perform these experiments for Face classifiers and Cats/Dogs. These datasets are chosen since they are more familiar to a layperson, making it more likely that users write similar words when describing an attribute.
We provide the full answers of users in the \href{http://explaining-in-style.github.io/supmat.html}{supplementary material}, yet a qualitative assessment of the responses leads to similar conclusions as in the first part of the study regarding the distinctness and coherence. For instance, all users used the word ``glasses" when describing the top coordinate extracted by \ours{} for the perceived age classifier. In general for all coordinates extracted by \ours{}, except one, there is a common word shared by all descriptions, and only in two coordinates the most common word is the same. On the other hand, for \sspacename{} less than half of the coordinates have a common word in all their descriptions, and two pairs of coordinates have the same most common word.
% \yossig{Yoav - elaborate? Amir suggested - "For example, the first attribute that was discovered for the perceived gender classifier was named A by X of Y users, as shown in figure B. Please refer to the supplemental material for all the results."} \\

% Adding $\mathcal{L}_{cls}$ does not change results significantly on most domains, yet as expected it is crucial for success in the Diabetic Retinopathy dataset. For this dataset, the classification loss imposes the model to include markers of the disease when generating images of unhealthy retinal fundus. When this loss is not included, all generated images appear healthy and therefore the classification simply cannot be flipped.
% Furthermore, observing \figref{fig:retina_top_5_attributes} it is strikingly clear that each of the top attributes extracted for this dataset corresponds to a different indicator of the disease (Diabetic Macular Edema).
%\subsection{\scalebox{0.9}[1]{`Sufficiency': Effect of Attributes on Classifier Output}}
\subsubsection{\scalebox{0.9}[1]{`Sufficiency': Effect of Attributes on Classifier Output}}
\label{sub:effect_on_classifier}
To test the effect of the attributes on the classifier, we ask if interventions on a small set of attributes can flip the classifier decision. Specifically, we try modifications on top $k$ attributes up-to $k=10$.
% , and try the $k$ modifications corresponding to changing only the first $i$ of these. 
We then measure the fraction of images that can be flipped (hence explained) in this way.
\begin{table}[th]
\begin{center}
\scalebox{0.9}{
% \begin{tabular}{|l||c|c|c|c|c|c|}
% \hline
% \textbf{} & \textbf{Perceived Gender} & \textbf{Perceived Age} & \textbf{Cats/Dogs} & \textbf{Wild Cats} & \textbf{Plants} & \textbf{Retina}  \\
% \hline
% % Wu \etal \cite{wu2020stylespace} & 44.8\% & 27.5\% & 1.5\% & ? & 10.6\% & 100\% \\
% Wu \etal \cite{wu2020stylespace} & 14.3\% & 16.9\% & 1.0\% & 11.8\% & 14.6\% & 0\% \\
% \hline
% {\ours} w/o classifier based training & 82.7\% & 93.0\% & 15.7\% & 18.9\% & 58.2\% &  0\% \\
% \hline
% % Ours, w/o conditioning & 82.1\% & 95.8\% & 26.6\% & 5.1\% & ? & 81.2\% \\
% % \hline
% {\ours}  & 83.2\% & 93.9\% & 25.0\% & 66.7\% & 91.2\% & 100\% \\
% \hline
% \end{tabular}
\begin{tabular}{l|c|c|c}
\toprule%\hline
\textbf{} &\textbf{ Wu \etal} & \textbf{Ours w/o CST} & \textbf{Ours}  \\
%\textbf{} & \cite{wu2020stylespace} & \textbf{based training} &   \\
\midrule%\hline
Perceived Gender 	& 14.3\% & 82.7\% & 83.2\%  \\
% 	&  &  & \\ 
%\hline
Perceived Age	& 16.9\% & 93.0\% & 93.9\%  \\
%\hline
Cats/Dogs 			& 1.0\% & 15.7\% & 25.0\% \\
%\hline
Wild Cats 			& 11.8\% & 18.9\% & 66.7\% \\
%\hline
Plants 				& 14.6\% & 58.2\% & 91.2\% \\
%\hline
Retina 				& 0.0\% & 0.0\% & 100\% \\
\bottomrule%\hline
\end{tabular}}
\end{center}
\caption{\textbf{Effect of Top-10 Attributes on the Classifier}. The fraction of images for which the classification flipped when modifying 
top-$k$ attributes up to $k=10$ 
%a subset of the top-10 attributes 
(see \secref{sub:effect_on_classifier}). Attributes discovered by {\ours} affect classification results for a much larger percentage of images than the baseline methods. On the face domains,  {\attproc} finds sufficient attributes even on standard StyleGAN2, while in other domains, classifier-specific training is required. On the Cats/Dogs classifier, due to the large visual differences between the two classes, top-10 attributes are not enough. 40 attributes are required to flip the classifier in $94\%$ of the images.}
% \caption{{\bf Comparison of the percent of images that flipped their class by the top-$10$ attributes}. Our method discovers attributes that have a larger effect on the classifier, as compared to \cite{wu2020stylespace}. Additionally removing the classifier based training results in performance degradation. For the Cats/Dogs classifier the top-40 chosen attributes flip the class of 94\% of the images.}
\label{table:ablation_sufficiency}
\vspace*{-0.2cm}

\end{table}

%For a single attribute, the effect on the classifier can be quantified by the Average Causal Effect \cite{pearl2009causality} (i.e. on average, how much does an intervention on the attribute change the output of the classifier). We adopt a similar metric to quantify the effect for a set attributes. Specifically, we consider the top-$k$ attributes, and calculate the fraction of images where modifying these attributes together results in the classifier switching its prediction. \amirg{not clear if this is subset change or something else}\yossig{ not a subset. a greedy search. } 
%(e.g., for binary prediction this means the softmax crosses the $0.5$ threshold after the change)

%that, when those attributes are changed in them, the classifier flips the prediction on them.
%We refer to a set of $k$ attributes as "sufficient", if for each image within a large set of  $N$ random images (e.g., $N=1000$), there exists a subset of these $k$ attributes that, when changed in the image, flips the classifier prediction on that image. To find such a sufficient set of attributes, our method accumulates the top sorted attributes until there exists such subset of attributes for each image among the set of $N$ images. 
\tabref{table:ablation_sufficiency} presents this measure on 1000 randomly chosen images. It can be seen that {\ours} achieves high explanation percentages on most domains.
%Additionally, we found that in some domains (e.g., retina), top-$3$ features also  
\comment{
The top-$10$ attributes allow our method to explain the decision of the classifiers for the vast majority of randomly selected images for almost all the domains. Namely, for almost each of these randomly tested images, there exists a subset of these top-10 attributes that flips the classifier's decision on it (e.g. from perceived female to perceived male, or vice versa). For example, in the retinopathy domain, each discovered attribute is a different indicator for the diabetic retinopathy (e.g. exudates, cotton wool and hemorrhages), as presented in \figref{fig:retina_top_5_attributes}. Therefore, a flip of the classifier decision for all the images can be achieved with fewer attributes (e.g. top-$3$). \yossig{Should we talk about the decrease in sufficiency in different domains?}
}
\tabref{table:ablation_sufficiency} also reports results of {\ours} without the classifier-specific training (i.e. without  conditional training and classifier loss). Note that this component has a dramatic effect on performance in the retina and plants domains. This is in line with the fact that the classes in these cases correspond to features that are more subtle and localized, thus less likely to be captured by a GAN that is oblivious to the classifier. Specifically, we verified that when training without classifier information, the generated images in the retina domain collapse to one class (``healthy'').
%\amirg{say we have more ablations in supp., assuming we do...}
%\yossig{Suggesting - Moreover, adding classifier loss without conditioning, flips $81\%$ of the retinal images, increasing the performance in comparison to removing fully the classifier base training.} \amirg{take some text from below}
\comment{
We further compared between the effect of the attributes on the classifier, when our method is trained with and without the classifier loss and the conditional input. As shown in \ref{table:ablation_sufficiency}, the top-$10$ attributes that are chosen by our method have bigger effect on the classifier when the generator is trained with classifier loss and conditioning. In several domains with unstructured fine-grained differences between classes (e.g. exudates on retinopathy images or bacterial spots on plant images) when the generator is trained without classification loss, it tends to generate image from one class only. Thus, both our extraction algorithm and \cite{wu2020stylespace} do not find  attributes that effect the classifier. For example, the retinopathy images that were generated by a GAN without classification loss were all classified by the class "healthy", and no subset of attributes that can flip the classifier prediction was found. Visual comparison of the chosen attributes appears in the supplemental material. \yossig{Explain why do we need a conditional input}
}
\comment{
\subsection{Effect on Classifier}
\label{sec:effect_on_classifier}

% More precisely, given a set of $M$ style coordinates $S$, we define their sufficiency as the percentage of images for which a change in some subset of $S$ flips the decision of the classifier. That is, we count the amount of images for which the set of attributes are sufficient to inflict a change in the class predicted by the classifier. \yoavw{Might be good to have a formal definition here, using the $\Delta[I, s, d]$ notation introduces earlier. Would be happy to hear what others think.}\\

The output of our method is a \textbf{sufficient} set of generative attributes.
\subsection{Diversity}
For each image, there might be more than one way to switch its prediction from class A to class B. For example, to make a person seem older, one can add wrinkles around either the mouth or the eyes, make their hair thinner or more white, or adjust the proportions of their face. This essentially provides alternative sufficient explanations for the classifier decision. These different explanations can be illustrated by visualizing different counterfactual variations for that specific image.

Our method can be used to find all the \textbf{diverse} subsets of attributes that, when applied, flip the label of a given image {\bf in a different way}. Figure \ref{fig:diversity} presents different sets of manipulations that flip the label of the input image from female to male. To find those minimal subsets, we enumerate over all the possible subsets of the sufficient top-$k$ attributes that were discovered by our method, and search for disjoint subsets that switch the classification result when each subset is applied to the image. Note that when $k$ is not very large, exhaustively checking all possible subsets of attributes is computationally feasible (e.g. for $k=10$, there exist only $1023$ possible subsets).  
This separates our methods from previous methods like GANalyze \cite{goetschalckx2019ganalyze} which only offer one way to change the label (as shown in figure \ref{fig:motivation}). StarGAN \cite{choi2020stargan} also offers multiple ways to change the label of a certain image, but these ways completely change the identity of the image, and so cannot be considered an explanation.
}
\comment{
\subsection{Human interpretability of the attributes}
Besides sufficiency and diversity, the ultimate goal of our method is to extract attributes that are semantic. The attributes we consider are extracted in an unsupervised manner, therefore they are not associated with any ground truth labels. Furthermore they may correspond to notions for which human annotations do not exist (e.g. in medical imaging). Therefore we design a user study to verify the interpretability of our attributes, and refrain from using labels found in common datasets.

The first part of our study examines a purely visual aspect of interpretability. It shows that humans can recognize modifications in an attribute extracted by \ours{} after observing a few examples of such modifications, and that they are able to distinguish between two different attributes. Following a similar study to \cite{yeh2020completeness}\footnote{Their study also asks users to identify attributes, though they do not use a generative model. Hence they show users patches of images to demonstrate attributes, instead of the animations we use here.}, we show users two animated GIFs that demonstrate a change in a certain extracted attribute from \ours{} on two different images. Then out of two candidate animations (one changing the same attribute that was demonstrated and the other changing a different attribute), users are asked to select the candidate that changes the demonstrated attribute.

In domains that are familiar to most humans, such as face images and cats vs. dogs, we may also expect to extract attributes that can be described verbally by humans. Hence in our second study we show users animated GIFs that demonstrate a modification in an attribute, and ask them to describe in 1-4 words what is the attribute they see changing in the image. They are also given the option to mark a checkbox saying that the animations do not change a single attribute which they can describe. We perform the user study on the top 6 attributes extracted from a perceived gender classifier using \ours{} with a limited number of users, so that their answers are easy to examine manually. Their full answers are given in the \href{http://explaining-in-style.github.io/supmat.html}{supplementary material}, and qualitative assessment of their answers suggests that most users give informative descriptions that are similar to one another.
}
\comment{
Since our attributes are extracted in an unsupervised manner, our method does not associate any labels to the attributes it extracts. In order for these attributes to be useful for explainability, they need to be interpretable by humans just from observing the change each image undergoes. This means that if a person sees a few examples for manipulation of the same attribute, we expect that this person could predict how would another image change according to this attribute. Even better, since understanding an idea means being able to describe it in words, a good attribute is one which a person could give literal meaning to.

To test the interpretability of our attributes, we designed two human studies. In one study, inspired by \cite{yeh2020completeness}, we presented workers two pairs of images which are manipulated along the same attribute... TODO yoav - complete.}
% \section{Experiments}

% \subsection{Ablation Study}
\label{sub:ablation}

\label{subsection:training_with_classifier}

\section{Conclusion}
%This work proposes a method for discovering and visualizing semantic attributes to explain a given classifier. 
We introduced a new technique
%for learning a classifier-specific StyleSpace by adding classifier based training to StyleGAN2. We used this StyleSpace to
for generating different counterfactual explanations for a given classifier on a given image. Our results demonstrate that these attributes correspond to clear visual concepts {\em and} directly affect classifier decisions.
%semantically meaningful, distinct and visually coherent for various domains. Thus the method can visually explain a given classifier in a human-interpretable way. 
We believe that {\ours} is a promising step towards detection and mitigation of previously unknown biases in classifiers. Additionally, our focus on multiple-attribute based counterfactuals is key to providing new insights about previously opaque classification processes and aiding in the process of scientific discovery. 
%\newpage
% Figures:

% - Teaser (high level setup + diversity of domains)

% - motivation - compared to heatmaps + ganalyzer

% - Architecture
% - Domains + their attributes (3 per domain), best attributes for single image (2-3 domains)

% - Different ways for changing an image

% - comparison to other methods (evaluations)

% - Human study setup + table

% - ablation study tables

% - sufficiency and diversity

{\small
\bibliographystyle{ieee_fullname}
\bibliography{egbib}

\begin{thebibliography}{10}\itemsep=-1pt

\bibitem{antoran2020getting}
Javier Antor{\'a}n, Umang Bhatt, Tameem Adel, Adrian Weller, and
  Jos{\'e}~Miguel Hern{\'a}ndez-Lobato.
\newblock Getting a clue: A method for explaining uncertainty estimates.
\newblock {\em arXiv preprint arXiv:2006.06848}, 2020.

\bibitem{choi2020stargan}
Yunjey Choi, Youngjung Uh, Jaejun Yoo, and Jung-Woo Ha.
\newblock Stargan v2: Diverse image synthesis for multiple domains.
\newblock In {\em Proceedings of the IEEE/CVF Conference on Computer Vision and
  Pattern Recognition}, pages 8188--8197, 2020.

\bibitem{collins2020editing}
Edo Collins, Raja Bala, Bob Price, and Sabine Susstrunk.
\newblock Editing in style: Uncovering the local semantics of gans.
\newblock In {\em Proceedings of the IEEE/CVF Conference on Computer Vision and
  Pattern Recognition}, pages 5771--5780, 2020.

\bibitem{dhamdhere2018important}
Kedar Dhamdhere, Mukund Sundararajan, and Qiqi Yan.
\newblock How important is a neuron?
\newblock {\em arXiv preprint arXiv:1805.12233}, 2018.

\bibitem{esser2020disentangling}
Patrick Esser, Robin Rombach, and Bjorn Ommer.
\newblock A disentangling invertible interpretation network for explaining
  latent representations.
\newblock In {\em Proceedings of the IEEE/CVF Conference on Computer Vision and
  Pattern Recognition}, pages 9223--9232, 2020.

\bibitem{ghorbani2019towards}
Amirata Ghorbani, James Wexler, James Zou, and Been Kim.
\newblock Towards automatic concept-based explanations.
\newblock {\em arXiv preprint arXiv:1902.03129}, 2019.

\bibitem{goetschalckx2019ganalyze}
Lore Goetschalckx, Alex Andonian, Aude Oliva, and Phillip Isola.
\newblock Ganalyze: Toward visual definitions of cognitive image properties.
\newblock In {\em Proceedings of the IEEE/CVF International Conference on
  Computer Vision}, pages 5744--5753, 2019.

\bibitem{GoodfellowGAN}
Ian~J. Goodfellow, Jean Pouget-Abadie, Mehdi Mirza, Bing Xu, David
  Warde-Farley, Sherjil Ozair, Aaron Courville, and Yoshua Bengio.
\newblock Generative adversarial nets.
\newblock In {\em Proceedings of the 27th International Conference on Neural
  Information Processing Systems - Volume 2}, NIPS'14, page 2672–2680,
  Cambridge, MA, USA, 2014. MIT Press.

\bibitem{goodfellow2014explaining}
Ian~J Goodfellow, Jonathon Shlens, and Christian Szegedy.
\newblock Explaining and harnessing adversarial examples.
\newblock {\em arXiv preprint arXiv:1412.6572}, 2014.

\bibitem{goyal2019explaining}
Yash Goyal, Amir Feder, Uri Shalit, and Been Kim.
\newblock Explaining classifiers with causal concept effect (cace).
\newblock {\em arXiv preprint arXiv:1907.07165}, 2019.

\bibitem{goyal2019counterfactual}
Yash Goyal, Ziyan Wu, Jan Ernst, Dhruv Batra, Devi Parikh, and Stefan Lee.
\newblock Counterfactual visual explanations.
\newblock In {\em International Conference on Machine Learning}, pages
  2376--2384. PMLR, 2019.

\bibitem{guo2019simple}
Chuan Guo, Jacob Gardner, Yurong You, Andrew~Gordon Wilson, and Kilian
  Weinberger.
\newblock Simple black-box adversarial attacks.
\newblock In {\em International Conference on Machine Learning}, pages
  2484--2493. PMLR, 2019.

\bibitem{Mobilenet}
Andrew~G. Howard, Menglong Zhu, Bo Chen, Dmitry Kalenichenko, Weijun Wang,
  Tobias Weyand, Marco Andreetto, and Hartwig Adam.
\newblock Mobilenets: Efficient convolutional neural networks for mobile vision
  applications.
\newblock {\em CoRR}, abs/1704.04861, 2017.

\bibitem{hughes2015open}
David Hughes, Marcel Salath{\'e}, et~al.
\newblock An open access repository of images on plant health to enable the
  development of mobile disease diagnostics.
\newblock {\em arXiv preprint arXiv:1511.08060}, 2015.

\bibitem{karras2019style}
Tero Karras, Samuli Laine, and Timo Aila.
\newblock A style-based generator architecture for generative adversarial
  networks.
\newblock In {\em Proceedings of the IEEE/CVF Conference on Computer Vision and
  Pattern Recognition}, pages 4401--4410, 2019.

\bibitem{karras2020analyzing}
Tero Karras, Samuli Laine, Miika Aittala, Janne Hellsten, Jaakko Lehtinen, and
  Timo Aila.
\newblock Analyzing and improving the image quality of {StyleGAN}.
\newblock In {\em Proceedings of the IEEE/CVF Conference on Computer Vision and
  Pattern Recognition}, pages 8110--8119, 2020.

\bibitem{kim2018interpretability}
Been Kim, Martin Wattenberg, Justin Gilmer, Carrie Cai, James Wexler, Fernanda
  Viegas, et~al.
\newblock Interpretability beyond feature attribution: Quantitative testing
  with concept activation vectors (tcav).
\newblock In {\em International conference on machine learning}, pages
  2668--2677. PMLR, 2018.

\bibitem{krause2018grader}
Jonathan Krause, Varun Gulshan, Ehsan Rahimy, Peter Karth, Kasumi Widner,
  Greg~S Corrado, Lily Peng, and Dale~R Webster.
\newblock Grader variability and the importance of reference standards for
  evaluating machine learning models for diabetic retinopathy.
\newblock {\em Ophthalmology}, 125(8):1264--1272, 2018.

\bibitem{mothilal2020explaining}
Ramaravind~K Mothilal, Amit Sharma, and Chenhao Tan.
\newblock Explaining machine learning classifiers through diverse
  counterfactual explanations.
\newblock In {\em Proceedings of the 2020 Conference on Fairness,
  Accountability, and Transparency}, pages 607--617, 2020.

\bibitem{narayanaswamy2020scientific}
Arunachalam Narayanaswamy, Subhashini Venugopalan, Dale~R Webster, Lily Peng,
  Greg~S Corrado, Paisan Ruamviboonsuk, Pinal Bavishi, Michael Brenner,
  Philip~C Nelson, and Avinash~V Varadarajan.
\newblock Scientific discovery by generating counterfactuals using image
  translation.
\newblock In {\em International Conference on Medical Image Computing and
  Computer-Assisted Intervention}, pages 273--283. Springer, 2020.

\bibitem{o2020generative}
Matthew O'Shaughnessy, Gregory Canal, Marissa Connor, Mark Davenport, and
  Christopher Rozell.
\newblock Generative causal explanations of black-box classifiers.
\newblock {\em arXiv preprint arXiv:2006.13913}, 2020.

\bibitem{pearl2009causality}
Judea Pearl.
\newblock {\em Causality}.
\newblock Cambridge university press, 2009.

\bibitem{rakhlin2018diabetic}
Alexander Rakhlin.
\newblock Diabetic retinopathy detection through integration of deep learning
  classification framework.
\newblock {\em bioRxiv}, page 225508, 2018.

\bibitem{rebuffi2020there}
Sylvestre-Alvise Rebuffi, Ruth Fong, Xu Ji, and Andrea Vedaldi.
\newblock There and back again: Revisiting backpropagation saliency methods.
\newblock In {\em Proceedings of the IEEE/CVF Conference on Computer Vision and
  Pattern Recognition}, pages 8839--8848, 2020.

\bibitem{selvaraju2017grad}
Ramprasaath~R Selvaraju, Michael Cogswell, Abhishek Das, Ramakrishna Vedantam,
  Devi Parikh, and Dhruv Batra.
\newblock Grad-cam: Visual explanations from deep networks via gradient-based
  localization.
\newblock In {\em Proceedings of the IEEE international conference on computer
  vision}, pages 618--626, 2017.

\bibitem{shen2020interpreting}
Yujun Shen, Jinjin Gu, Xiaoou Tang, and Bolei Zhou.
\newblock Interpreting the latent space of {GANs} for semantic face editing.
\newblock In {\em Proceedings of the IEEE/CVF Conference on Computer Vision and
  Pattern Recognition}, pages 9243--9252, 2020.

\bibitem{shen2020interfacegan}
Yujun Shen, Ceyuan Yang, Xiaoou Tang, and Bolei Zhou.
\newblock Interfacegan: Interpreting the disentangled face representation
  learned by {GANs}.
\newblock {\em IEEE Transactions on Pattern Analysis and Machine Intelligence},
  2020.

\bibitem{shen2020closed}
Yujun Shen and Bolei Zhou.
\newblock Closed-form factorization of latent semantics in {GANs}.
\newblock {\em arXiv preprint arXiv:2007.06600}, 2020.

\bibitem{shocher2020semantic}
Assaf Shocher, Yossi Gandelsman, Inbar Mosseri, Michal Yarom, Michal Irani,
  William~T Freeman, and Tali Dekel.
\newblock Semantic pyramid for image generation.
\newblock In {\em Proceedings of the IEEE/CVF Conference on Computer Vision and
  Pattern Recognition}, pages 7457--7466, 2020.

\bibitem{shrikumar2017learning}
Avanti Shrikumar, Peyton Greenside, and Anshul Kundaje.
\newblock Learning important features through propagating activation
  differences.
\newblock In {\em International Conference on Machine Learning}, pages
  3145--3153. PMLR, 2017.

\bibitem{singla2019explanation}
Sumedha Singla, Brian Pollack, Junxiang Chen, and Kayhan Batmanghelich.
\newblock Explanation by progressive exaggeration.
\newblock {\em arXiv preprint arXiv:1911.00483}, 2019.

\bibitem{singla2021explaining}
Sumedha Singla, Brian Pollack, Stephen Wallace, and Kayhan Batmanghelich.
\newblock Explaining the black-box smoothly-a counterfactual approach.
\newblock {\em arXiv preprint arXiv:2101.04230}, 2021.

\bibitem{wachter2017counterfactual}
Sandra Wachter, Brent Mittelstadt, and Chris Russell.
\newblock Counterfactual explanations without opening the black box: Automated
  decisions and the gdpr.
\newblock {\em Harv. JL \& Tech.}, 31:841, 2017.

\bibitem{wah2011caltech}
Catherine Wah, Steve Branson, Peter Welinder, Pietro Perona, and Serge
  Belongie.
\newblock The caltech-ucsd birds-200-2011 dataset.
\newblock 2011.

\bibitem{wu2020stylespace}
Zongze Wu, Dani Lischinski, and Eli Shechtman.
\newblock Stylespace analysis: Disentangled controls for {StyleGAN} image
  generation.
\newblock {\em arXiv preprint arXiv:2011.12799}, 2020.

\bibitem{xu2020attribution}
Shawn Xu, Subhashini Venugopalan, and Mukund Sundararajan.
\newblock Attribution in scale and space.
\newblock In {\em Proceedings of the IEEE/CVF Conference on Computer Vision and
  Pattern Recognition}, pages 9680--9689, 2020.

\bibitem{Yinghao2020}
Yinghao Xu, Yujun Shen, Jiapeng Zhu, Ceyuan Yang, and Bolei Zhou.
\newblock Generative hierarchical features from synthesizing images.
\newblock {\em CoRR}, abs/2007.10379, 2020.

\bibitem{yeh2020completeness}
Chih-Kuan Yeh, Been Kim, Sercan Arik, Chun-Liang Li, Pradeep Ravikumar, and
  Tomas Pfister.
\newblock On completeness-aware concept-based explanations in deep neural
  networks.
\newblock 2020.

\bibitem{zhang2018perceptual}
Richard Zhang, Phillip Isola, Alexei~A Efros, Eli Shechtman, and Oliver Wang.
\newblock The unreasonable effectiveness of deep features as a perceptual
  metric.
\newblock In {\em CVPR}, 2018.

\end{thebibliography}
}

% \newpage
% \input{appendix.tex}

% \newpage
% \begin{center}
% \textbf{\large Rebuttal}
% \end{center}
% \input{rebuttal/contents.tex}
\end{document}